\documentclass[runningheads]{llncs}
\usepackage[final]{eccv}

\usepackage{graphicx}
\usepackage{eccvabbr}

\usepackage{tikz}
\usepackage{comment}
\usepackage{amsmath,amssymb} 
\usepackage{color}
\usepackage{subcaption} 

\usepackage[accsupp]{axessibility}  

\usepackage[width=122mm,left=12mm,paperwidth=146mm,height=193mm,top=12mm,paperheight=217mm]{geometry}

\definecolor{cvprblue}{rgb}{0.21,0.49,0.74}

\newcommand{\sbest}[1]{\underline{#1}} 

\usepackage[pagebackref,breaklinks,colorlinks,citecolor=cvprblue]{hyperref}
\usepackage{diagbox}
\usepackage{multirow}
\usepackage{dsfont}
	
\usepackage{color, colortbl}
\definecolor{lightgray}{gray}{0.9}

\newcommand\blfootnote[1]{%
  \begingroup
  \renewcommand\thefootnote{}\footnote{#1}%
  \addtocounter{footnote}{-1}%
  \endgroup
}

\usepackage{xcolor}
\usepackage{multirow}
\usepackage{xcolor}
\usepackage{float}
\usepackage{capt-of}
\usepackage{booktabs}
\usepackage{varwidth}
\usepackage{tablefootnote}
\usepackage{multirow}
\usepackage{multicol}
\usepackage{siunitx}
\usepackage{tablefootnote}
\usepackage{pdflscape}
\usepackage{pifont}
\usepackage{subcaption} 
\usepackage{booktabs}
\usepackage{makecell}
\usepackage{diagbox}
\usepackage{amsmath}
\usepackage{dsfont}
\definecolor{columbiablue}{rgb}{0.61, 0.87, 1.0}
\definecolor{bisque}{rgb}{1.0, 0.89, 0.77}

\begin{document}
\pagestyle{headings}
\mainmatter

\title{$3\times2$: 3D Object Part Segmentation by 2D Semantic Correspondences} 


\titlerunning{3-By-2}
%
\author{Anh Thai\inst{1,2} \and Weiyao Wang\inst{2} \and Hao Tang\inst{2} \and Stefan Stojanov\inst{1} \and James M. Rehg\inst{3} \and Matt Feiszli\inst{2}}
\authorrunning{A. Thai et al.}
%
\institute{Georgia Institute of Technology \and
Meta AI, FAIR\and University of Illinois Urbana-Champaign}

\maketitle
\blfootnote{\footnotesize{Work done as an intern at Meta AI (FAIR).}}
\begin{center}
    \centering
    \captionsetup{type=figure}
    \vspace{-5mm}
    \includegraphics[width=\textwidth]{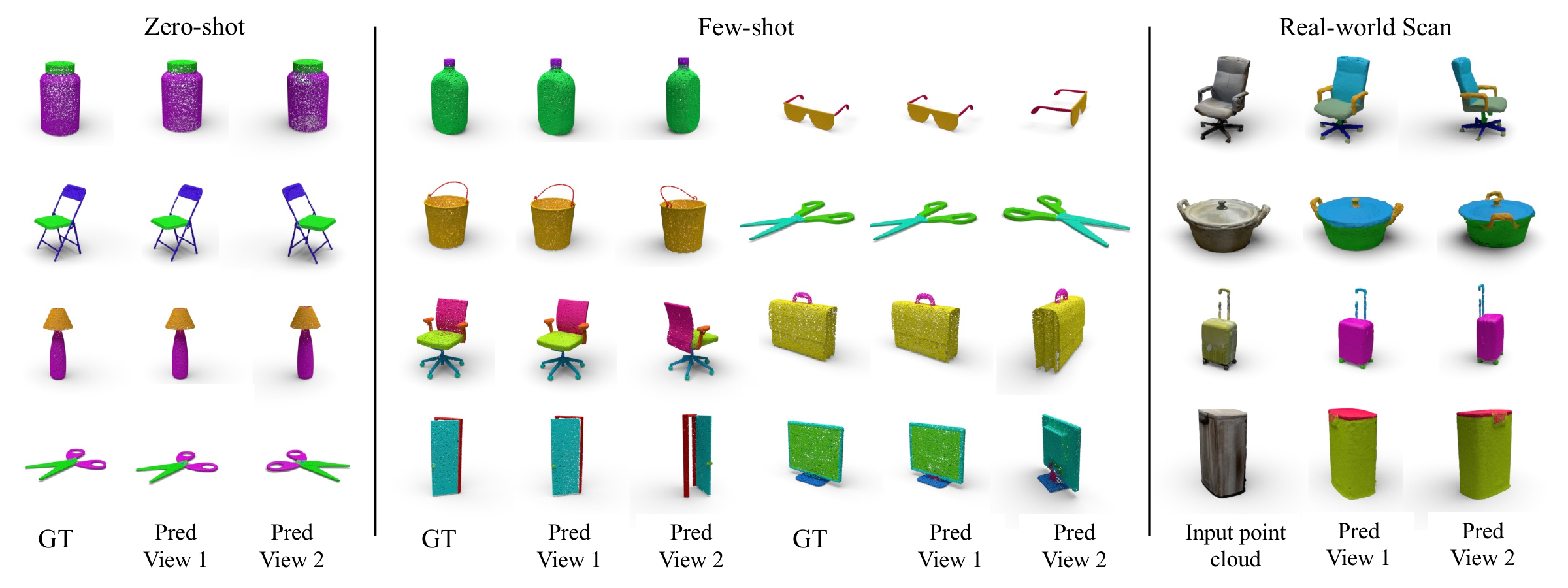}
    \vspace{-10pt}
    
    \captionof{figure}{We propose \textbf{3-By-2}, a novel training-free method for low-shot 3D object part segmentation that achieves SOTA performance on both zero-shot and few-shot settings.}
	\label{fig:teaser}
    \vspace{-10pt}
\end{center}

\begin{abstract}
3D object part segmentation is essential in computer vision applications. While substantial progress has been made in 2D object part segmentation, the 3D counterpart has received less attention, in part due to the scarcity of annotated 3D datasets, which are expensive to collect. In this work, we propose to leverage a few annotated 3D shapes or richly annotated 2D datasets to perform 3D object part segmentation. We present our novel approach, termed 3-By-2 that achieves SOTA performance on different benchmarks with various granularity levels. By using features from pretrained foundation models and exploiting semantic and geometric correspondences, we are able to overcome the challenges of limited 3D annotations. Our approach leverages available 2D labels, enabling effective 3D object part segmentation. Our method 3-By-2 can accommodate various part taxonomies and granularities, demonstrating interesting part label transfer ability across different object categories. Project website: \url{https://ngailapdi.github.io/projects/3by2/}.
\end{abstract}
\section{Introduction}
\label{sec:intro}

3D object part understanding is essential in various research fields and applications, such as robotics \cite{varadarajan2011object,nadeau2023sum,liu2023composable} and graphics~\cite{Kalogerakis:2010:labelMeshes}. Through our understanding of the world, objects can be decomposed into parts based on diverse properties (e.g., geometry or affordance~\cite{liu2022autogpart,deng20213d}). However, these different decompositions do not always align with one another---the same object can be segmented into parts differently depending on the specific use case. For instance, a driver might perceive a car in terms of its functional components like the steering wheel, accelerator pedal, and brake pedal. Conversely, a manufacturing worker may view the car as an assembly of structural parts, such as the frame, bumper, and windshield. Further, various parts with similar functionalities or structures can be shared among different object classes (e.g., the term ``leg" can apply to multiple furniture items). How can we design a 3D part segmentation system that has high performance across such different requirements and scenarios?

Recent works in 3D part segmentation have integrated language as an additional input~\cite{liu2023partslip,abdelreheem2023satr,zhu2023pointclip} by leveraging vision-language models to prompt the segmentation. However, grounding visual parts using language is inherently ambiguous.  This is because parts can be described using diverse phrases that may include synonyms, various levels of detail, and differences in terminology (structural vs functional), which presents challenges for these models~\cite{liu2023partslip}. 
In contrast, images capture rich information about object shapes, textures and spatial part relationships. These properties can directly be parsed and compared using visual similarities between objects despite differences in linguistic expression. Therefore, it is important to study the limits and potentials of reasoning about visual similarity for generalization across different objects and categories. 

In this work, we investigate the 3D part segmentation task from this different perspective and propose a novel method called \textbf{3-By-2}. Since labeling 3D data is expensive, we design 3-by-2 to leverage existing extensively annotated 2D part segmentation datasets~\cite{ramanathan2023paco,he2022partimagenet} or a few-labeled 3D shapes to perform object part segmentation \emph{without additional training or finetuning}. Our method does not need any language input and can flexibly handle segmentation tasks at various levels of granularity. 

We build our method based on the observation that because objects are constructed from parts, and because various objects often share a common set of parts with similar visual structures, this should allow part label transfer from one object to another without any language description. Recent studies~\cite{tang2023emergent,zhang2023tale} have demonstrated the strong 2D semantic correspondences encoded by features of image diffusion models that generalize across different domains (e.g. sketch vs real images). To label a query 3D object point cloud, we leverage these strong representations to perform 2D pixel correspondence-based label transfer from in-the-wild 2D datasets or 2D renders of a few labeled 3D objects. To the best of our knowledge, we are the first to use diffusion model features for semantic label transfer in the context of 3D part segmentation.


\begin{figure*}[t]
\centering
	\includegraphics[width=\linewidth]{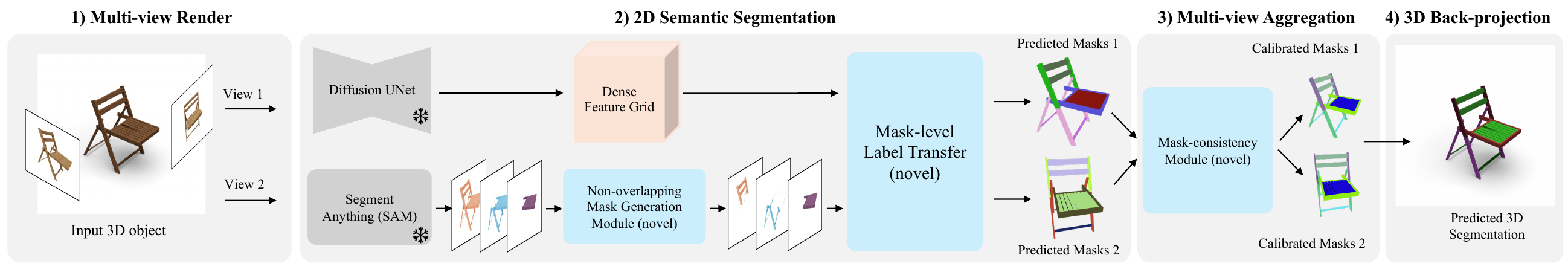}
   \vspace{-15pt}
	\caption{\textbf{Overview of our proposed method 3-By-2.} (1) Render the input object in multiple camera viewpoints, (2) Perform 2D part segmentation on each view individually by leveraging 2D semantic correspondences and 2D class-agnostic segmentation model, (3) Aggregate the 2D predictions from multiple views using our proposed mask-consistency module, (4) Back-project the predictions to 3D using depth information.}
	\label{fig:pipeline_overview}
  \vspace{-15pt}
\end{figure*}

While it might seem that obtaining 2D part labels for multi-view renders of an object through label transfer and back-projection into 3D is intuitively straightforward, a high performance and efficient implementation requires careful consideration of the challenges of 3D part segmentation: 1) Precise determination of 3D object part boundaries, which is particularly challenging for unstructured data like point clouds, and 2) Flexible adaptation to different levels of part granularity.  To this end, we introduce three novel elements of our method: non-overlapping generation, mask-level label transfer and mask-consistency modules (see Fig.~\ref{fig:pipeline_overview}). These components work efficiently together to ensure precise 3D part segmentation masks and boundaries across a range of object categories and part levels (Fig.~\ref{fig:teaser} and Tables~\ref{tab:partnete_few},~\ref{tab:partnete_zs},~\ref{tab:partnet}).

Overall, 3-By-2 is a training-free method \emph{independent} of language inputs, instead relying solely on the 2D labels provided by a 2D database. Unlike previous methods that require 3D segmentation priors like point-cloud clusters~\cite{liu2023partslip} or mesh surface information~\cite{abdelreheem2023satr,sharma2022mvdecor}, our approach has only a single requirement: calibrated cameras for back-projection. This can be known during the rendering process or predicted using SfM approaches. 



We validate the performance of our approach with PartNet-Ensembled~\cite{liu2023partslip}, a dataset tailored for language-input models, and PartNet~\cite{mo2019partnet}, which is not tailored for language. These datasets exhibit multiple levels of granularity. Notably, unlike previous approaches that require category-specific fine-tuning for few-shot scenarios~\cite{liu2023partslip,sharma2022mvdecor}, 3-By-2 achieves SOTA performance without any training or fine-tuning requirements in either a zero-shot or few-shot setting. Additionally, we identify that models with language inputs exhibit suboptimal performance with highly fine-grained part terminologies. This highlights the advantages of our approach, which effectively handles these fine-grained object parts. Furthermore, we conduct comprehensive ablation studies and demonstrate the transferability of parts across different object categories, which benefits the understanding of object part compositionality.

In summary, our contributions are 4-fold: 
\begin{itemize}
    \item A novel, training-free method, 3-By-2, that achieves SOTA performance on benchmarks with different levels of granularity for zero-shot and few-shot 3D object part segmentation.
    \item The first to provide an effective approach for leveraging image diffusion model's features~\cite{tang2023emergent} to establish 2D semantic correspondences in the context of 3D part segmentation.
    \item Novel non-overlapping mask generation, mask-level label transfer, and mask-consistency modules that effectively transfer part labels from 2D database and extrapolate them to 3D.
    \item Demonstrating the flexibility of 3-By-2 in accommodating various database settings and in generalizing between different object categories.
\end{itemize}


\section{Related Work}
\label{sec:related_work}
\subsection{3D Part Segmentation}




In contrast to its 2D counterpart, the progress in this field has been relatively limited, primarily due to the high cost associated with collecting and annotating 3D datasets. Currently, all of the available large-scale annotated 3D object part datasets are synthetic~\cite{mo2019partnet,Yi16,wang2022ikea,li20223dcompat}. The most widely used benchmarks~\cite{mo2019partnet,Yi16} are predominantly derived from objects within the ShapeNetCore~\cite{chang2015shapenet} dataset. This problem has been tackled using architectures that take 3D representations~\cite{mo2019partnet,qian2022pointnext} as inputs. These methods were trained in a supervised manner, requiring large-scale annotated data. More recent approaches have attempted to investigate data-efficient training scenarios where only a few 3D shapes are annotated~\cite{liu2023partslip,Wang_2020_CVPR,sharma2022mvdecor,zhou2023partslip++}. 



\subsection{Multi-view 2D-3D Segmentation Using Foundation Models}
Although multi-view approaches have been widely utilized in the past for 3D segmentation~\cite{dai20183dmv,Jaritz2019MultiViewPF,Zhao2021SimilarityAwareFN}, the rapid advancement of 2D foundation models~\cite{li2022grounded,kirillov2023segment} has encouraged more SOTA research aimed at leveraging these models to perform 3D segmentation in a multi-view fashion. CLIP~\cite{radford2021learning} and GLIP~\cite{li2022grounded} have been employed to integrate language information from multiple 2D views into 3D for open-vocabulary segmentation~\cite{peng2023openscene,takmaz2023openmask3d,liu2023partslip,zhou2023partslip++,abdelreheem2023satr}. SAM~\cite{kirillov2023segment}, due to its ability to output per-pixel masks, has been used as an effective tool for multi-view 2D-3D segmentation, both on 3D structures like point clouds~\cite{zhou2023partslip++,yang2023sam3d,takmaz2023openmask3d,Yu2023When3B,xu2023sampro3d} or in NeRF-style~\cite{cen2024segment}.

\noindent\textbf{Scene Segmentation.} Various combinations of foundation models have been explored for this task. While~\cite{takmaz2023openmask3d} leverages CLIP and SAM to support open-vocabulary 3D part segmentation, others use SAM with carefully designed prompts \cite{cen2024segment} or post-processing techniques~\cite{yang2023sam3d}. Building upon these successes, concurrent works~\cite{xu2023sampro3d,nguyen2023open3dis,huang2023segment3d} seek to improve SAM utilization strategies. Our work differs by focusing on part segmentation, which requires finer granularity. This distinction in objectives directly influences the processing of SAM predictions, tailored to suit their specific characteristics. For example, while scene segmentation methods may disregard or merge masks covering parts of objects, part segmentation approaches might encourage splitting, depending on the desired level of detail.

\noindent\textbf{Part Segmentation.} PartSLIP~\cite{liu2023partslip} and SATR~\cite{abdelreheem2023satr} were among the first to employ foundation models for this task, pioneering the use of GLIP for open-vocabulary segmentation. Concurrent works have seen the integration of SAM into their pipelines~\cite{zhou2023partslip++,xue2023zerops,kim2024partstad}. Zhou et al.~\cite{zhou2023partslip++} and Kim et al.~\cite{kim2024partstad} use SAM with GLIP-predicted bounding boxes, while Xue et al.~\cite{xue2023zerops} employ SAM with furthest point sampling for each view, extending predictions to 3D with GLIP labels. Our approach shares with these works the use of SAM for 2D segmentation before 3D aggregation. In contrast, our method focuses solely on visual cues without language inputs, employing image diffusion model's features~\cite{tang2023emergent}. To improve SAM's accuracy, we introduce a novel non-overlapping mask generation module, eliminating the need for GLIP-generated bounding boxes.

\subsection{Part Label Transfer using Correspondences}
Transferring labels from annotated datasets to non-annotated datasets has been considered recently in~\cite{sun2023going} for open-vocabulary 2D part segmentation and previously in~\cite{10.1016/j.cagd.2019.04.009,chen2020unsupervised} for 3D part segmentation. While~\cite{sun2023going} used DINOv1~\cite{caron2021emerging} feature representations for dense label transfer between related objects in the base classes and novel object classes, Zhu et al.~\cite{10.1016/j.cagd.2019.04.009} relied on classical SIFT~\cite{lowe2004distinctive} features for establishing correspondences in 2D images. Chen et al.~\cite{chen2020unsupervised}, in contrast, train a network to regress the correspondences directly on the input point cloud. 

We share with these approaches the use of semantic correspondences to identify optimal candidates for label transfer. However, our primary objective sets us apart significantly from~\cite{sun2023going}, as we focus on segmenting 3D objects. Compared to~\cite{10.1016/j.cagd.2019.04.009}, we leverage class-agnostic segmentation models to avoid dense pixel/patch sampling. Furthermore, unlike~\cite{chen2020unsupervised}, we do not require direct operations on 3D point clouds or any specific 3D representations. Additionally, we introduce a mask-consistency module for per mask label voting, rather than relying solely on small local patches.

\noindent\textbf{Semantic Correspondences from Foundation Models}. Many vision foundation models have demonstrated an inherent capability to implicitly capture semantic correspondences across different instances within the same category (e.g., matching chair backs) and across diverse categories (e.g., aligning dog's legs with cat's legs)~\cite{tang2023emergent,zhang2023tale,hedlin2023unsupervised,amir2021deep}. In this work, we leverage semantic correspondences established by~\cite{tang2023emergent} to transfer part labels from annotated 2D datasets to query 3D objects.
\section{Method}

Given a database $\mathcal{D}$ consisting of 2D part annotations, our goal is to segment each query object $q$ into parts using the visual part vocabulary provided by $\mathcal{D}$. Note that $\mathcal{D}$ can either be gathered from 2D (image) part datasets or from renders of a few 3D objects captured at different view-points. Our method consists of three main steps (Fig.~\ref{fig:pipeline_overview}): (1) render a set of 2D RGB images $\mathcal{I}q$ of 3D object $q$ from $K$ distinct camera viewpoints; (2) perform 2D part segmentation on the rendered images; (3) aggregate image-level predictions through a mask-consistency aggregation module to obtain 3D predictions.




\begin{figure*}[t]
\centering
	\includegraphics[width=0.95\linewidth]{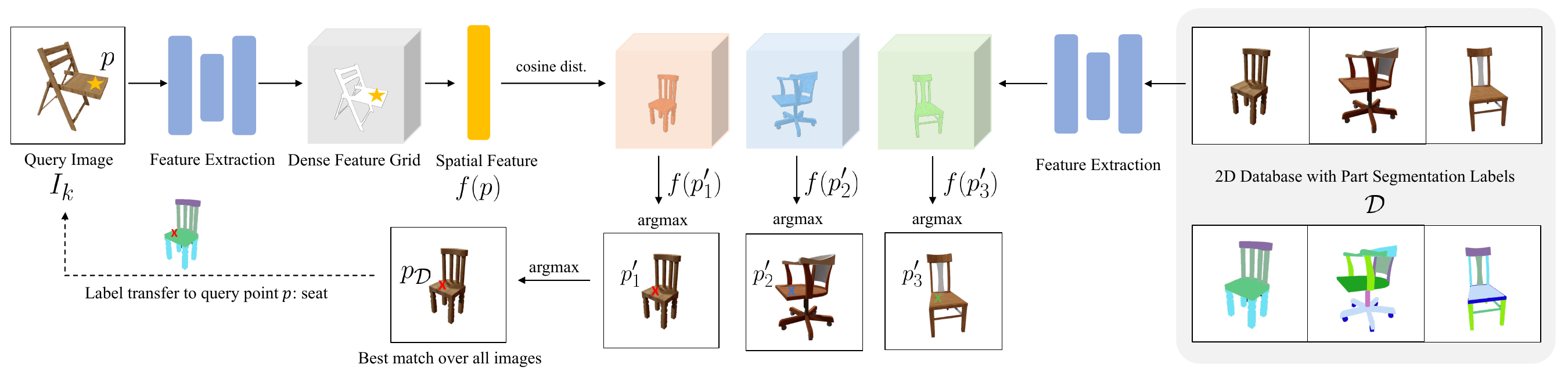}
  \vspace{-10pt}
 
	\caption{\textbf{The process of pixel-level part label transferring.} For each pixel $p$ in the query image $I_k$, we perform the following: (1) Extract the feature $f(p)$, along with the feature grid for each image $I_\mathcal{D}$ in the database $\mathcal{D}$, (2) Measure cosine similarity between $f(p)$ and the feature of each pixel within each feature grid, (3) Obtain the best match of $p$ over $\mathcal{D}$ by determining the most similar pixel $p_\mathcal{D}$ over all images $I_\mathcal{D}$, (4) Assign the label of $p$ is to be the label of $p_\mathcal{D}$.
 }
	\label{fig:label_transfer}
  \vspace{-15pt}
\end{figure*}

\begin{figure}[t]
\begin{subfigure}{0.5\textwidth}
\centering
	\includegraphics[width=0.9\linewidth]{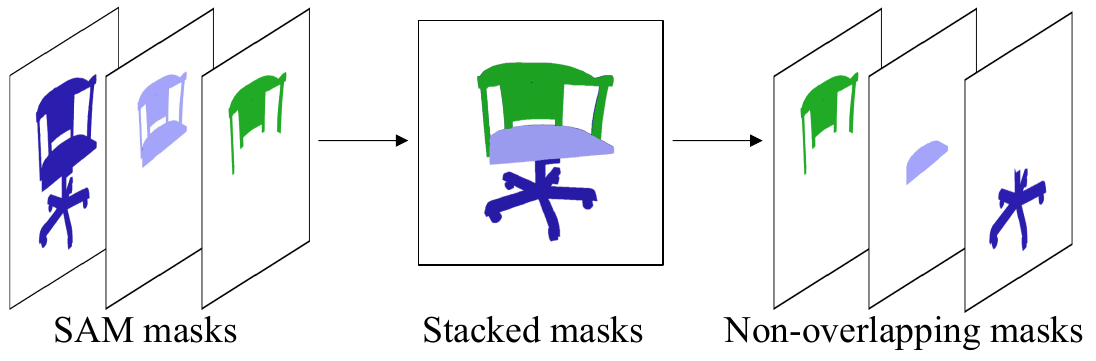}

	\label{fig:non_overlapping}
   \caption{\label{fig:non_overlapping}}
\end{subfigure}
  \vspace{-5pt}
\hspace{4pt}
\begin{subfigure}{0.5\textwidth}
\centering
	\includegraphics[width=0.95\linewidth]{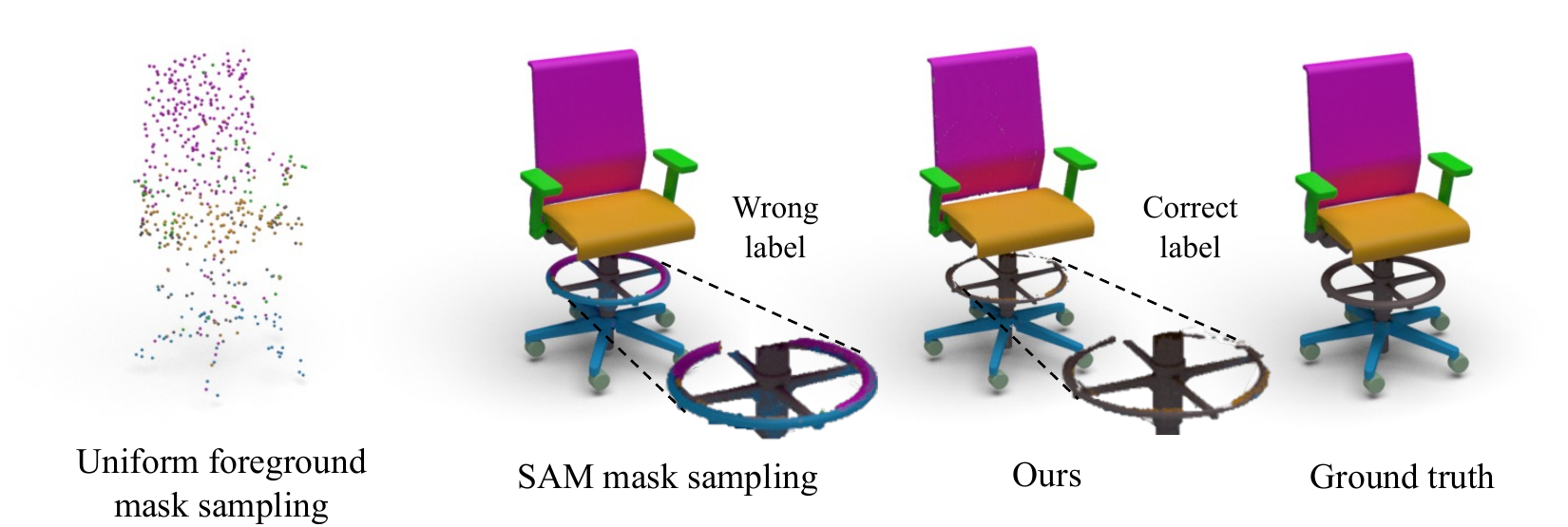}

	\label{fig:ablation}
  \caption{\label{fig:ablation}}
 
\end{subfigure}
  \vspace{-10pt}
	\caption{\textbf{(\subref{fig:non_overlapping}) Non-overlapping 2D Mask Proposal.} We address the issue of overlapping masks produced by SAM. The masks are first sorted by their areas. Subsequently, the smaller masks are stacked on top of the larger ones. Non-overlapping masks are obtained by taking the visible segment of each mask. \textbf{(\subref{fig:ablation}) Different mask sampling strategies for label transfer.} Our strategy provides accurate, dense prediction with clear part boundaries.}
   \vspace{-17pt}
\end{figure}

\subsection{2D Part Segmentation}
There are two primary approaches to tackle this task: (1) Top-down, using segmentation mechanisms such as SAM~\cite{kirillov2023segment}, or (2) Bottom-up, which involves labeling each pixel individually. While SAM produces high-quality 2D masks with sharp boundaries, it operates in a class-agnostic manner, often leading to high overlap between sub-parts, parts, and instances. Simply selecting the mask with the highest score may result in incorrect granularity and lacks the flexibility required for part segmentation. 

Conversely, doing label transfer for each pixel individually in the image is computationally impractical, particularly for part segmentation tasks where high resolution is preferred. Sparsely sampling and labeling pixels can result in under-segmented masks, particularly for smaller parts that are less likely to be sampled compared to larger parts (see Fig.~\ref{fig:ablation}). Moreover, accurately determining part boundaries for individual pixels can be challenging, which may result in increased errors when extrapolating to 3D, particularly with unstructured 3D representations like point clouds. These issues raise the important question: how do we transfer part labels and preserve part boundaries without sacrificing computational resources? 

To address this question, we propose a 2D segmentation method that combines the strengths of both approaches which consists of 3 novel components: (1) Single-pixel 2D label transfer using semantic correspondences derived from DIFT~\cite{tang2023emergent}, (2) Non-overlapping 2D mask proposal module, which refines SAM's multi-granularity predicted masks into non-overlapping part masks, and (3) Mask-level label transfer by integrating (1) and (2).

\noindent\textbf{Single-pixel 2D Label Transfer}. At the core of our method is the 2D label transfer process. The goal is to transfer pixel labels from the annotated 2D database $\mathcal{D}$ to the query RGB image $I_k\in\mathcal{I}q$: for a pixel $p$ in the foreground object in $I_k$, we aim to identify the best-matched pixel $p'$ in each image $I_\mathcal{D}$ in the database $\mathcal{D}$ and assign initial label to $p$ by $p'$. To this end, we leverage the established semantic correspondence of DIFT~\cite{tang2023emergent}. While recent works have demonstrated the effectiveness of image diffusion models in extracting semantic correspondences, as evidenced by evaluations on datasets like SPair-71K~\cite{min2019spair}, we are the first to leverage these features for transferring semantic labels in the context of 3D part segmentation. Specifically, $p'=\arg\max_{p'\in I_\mathcal{D}} \cos(f(p), f(p'))$ where $\cos$ and $f(x)$ denotes the cosine similarity score and the feature representing pixel $x$. The best pixel correspondence $p_\mathcal{D}$ of $p$ over the entire database is obtained by taking the most similar match within all the images in the database. Formally, $p_\mathcal{D}=\arg\max_\mathcal{D}p'$.
The label of $p$ is then assigned to be the label of $p_\mathcal{D}$ (see Fig.~\ref{fig:label_transfer}). 

\begin{figure}[t]
\centering
	\includegraphics[width=0.9\linewidth]{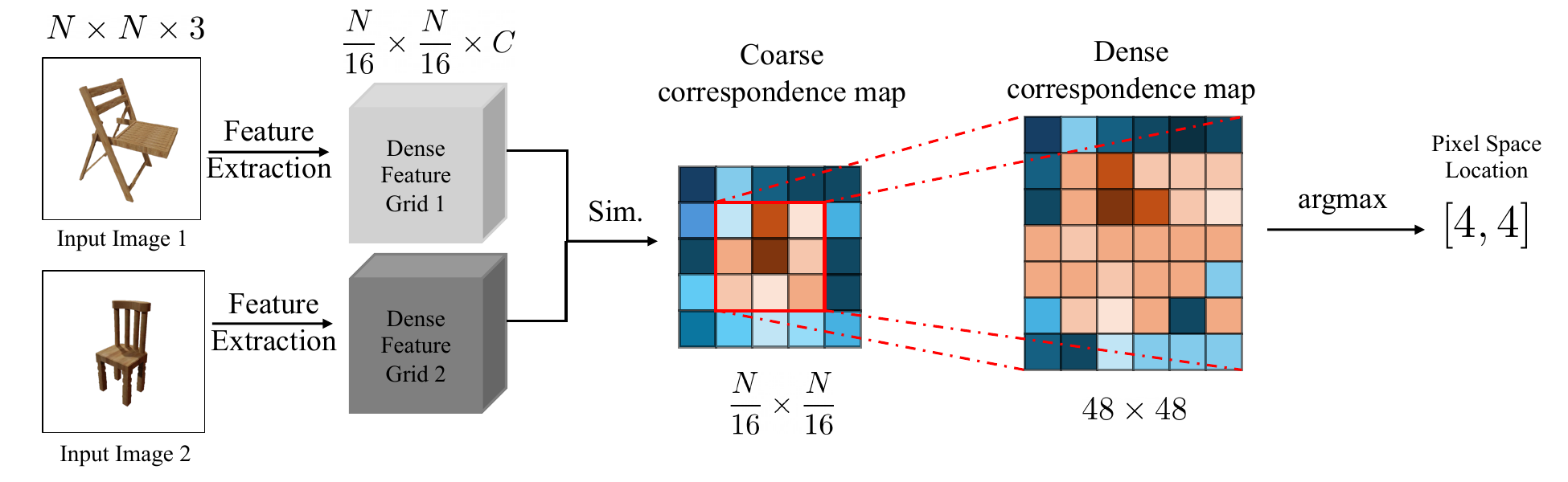}
  \vspace{-10pt}
	\caption{\textbf{Coase-to-fine correspondence search.} We first conduct searching on a coarse level to identify the region of best match. We then extract the $3\times 3$ window centered at this region in feature space for a fine search. This approach is approximately 2000 times faster in terms of wall time for large $N$ ($800\times 800$).}
	\label{fig:coarse_to_fine}
 \vspace{-20pt}
 
\end{figure}

\noindent\textit{\underline{Coarse-to-fine correspondence search}}. Finding the nearest neighbor for a query pixel across the entire database can be prohitively costly, especially for part segmentation which operates in high resolutions. We propose a coarse-to-fine strategy:
using the coarse feature maps generated by DIFT~\cite{tang2023emergent}, we first conduct the search at the coarse level to localize the region of the best match. We then extract the $3\times3$ window centered at this region (in feature space) for a fine search (see Fig.~\ref{fig:coarse_to_fine}). This approach ensures that we compute per-pixel similarity scores only within the region of interest, rather than across the entire image, improving computational efficiency. For instance, when processing a pair of images with a resolution of $800\times800$, coarse-to-fine correspondence search achieves a speed improvement of approximately 2000 times in terms of wall time.

\begin{figure}[t]
\begin{minipage}{0.5\textwidth}

\centering
	\includegraphics[width=0.95\linewidth]{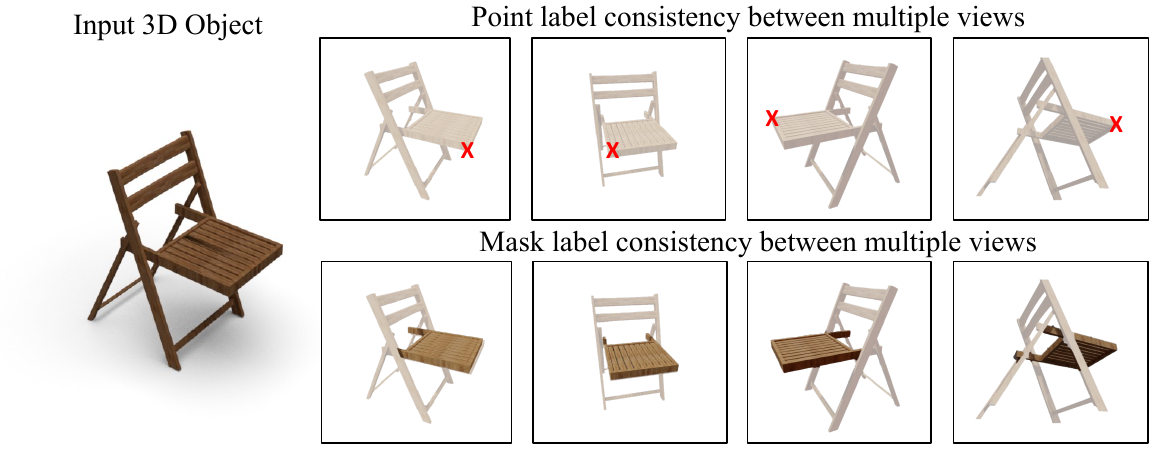}
	\caption{\textbf{Two approaches to aggregate 3D part labels from multiple 2D views}. Aggregating 3D part labels from multiple 2D views through geometric correspondence can be achieved by either point or mask label consistency. }
	\label{fig:mask_consistency}
 \vspace{-20pt}
\end{minipage}
\hspace{4pt}
\begin{minipage}{0.5\textwidth}
\centering
	\includegraphics[width=0.95\linewidth]{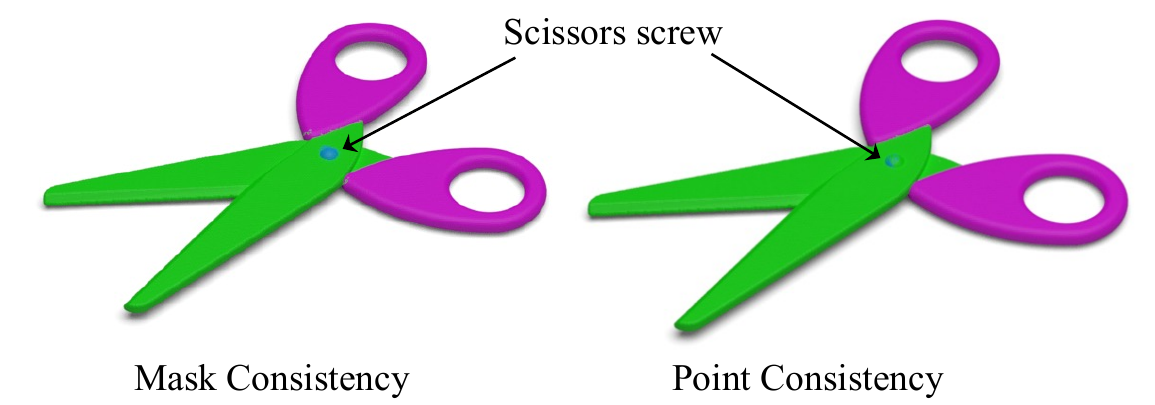}
	\caption{\textbf{Effectiveness of mask label consistency.} Enforcing consistency at the mask level can mitigate discrepancies at each individual point and contributes to smoother segmentation.}
	\label{fig:mc_viz}
 \vspace{-15pt}
 
\end{minipage}
\end{figure}

\noindent\textbf{Non-overlapping 2D Mask Proposal}. We propose the use of class-agnostic 2D part mask proposal, specifically from SAM~\cite{kirillov2023segment}. By assuming that each mask proposal corresponds to a subset of a part, we can then selectively sample pixels within each mask proposal for label transferring. The labels are subsequently propagated to each pixel of the 2D masks through a majority voting process based on the sampled pixels within the mask. 
To address the issue posed by the highly overlapping predictions from SAM's multi-granularity model, we introduce a non-overlapping 2D mask generation module. This module takes SAM masks as inputs and outputs a set of mutually exclusive 2D masks.

We arrange the SAM output masks in descending order of mask area and stack smaller masks on top of larger ones. This ensures that if mask $A$ is a subset of mask $B$, stacking $A$ on top of $B$ results in non-overlapping masks, namely $A$ and $B \setminus A$. Non-overlapping masks are finally obtained by taking the visible segments of each mask (see Fig.~\ref{fig:non_overlapping}).

\noindent\textbf{2D Mask Label Assignment}. After obtaining the non-overlapping masks, we sparsely sample pixels in each mask to transfer label. We then perform majority voting to assign the dominant label for each 2D mask, weighted by the confidence score (cosine similarity) of the best pixel correspondence matches.

\subsection{Mask-consistency Aggregation Module}
Given a set of 2D RGB images with part segmentation predictions, we aim to extrapolate these segmentation labels to 3D using geometric correspondences. Prior works~\cite{liu2023partslip,sharma2022mvdecor} aggregate multi-view information for each 3D point or mesh triangle face through a weighted sum of multi-view 2D predictions. To fully maintain the high-quality part boundaries predicted by SAM in 2D, we choose to aggregate multi-view predictions for each 2D mask instead. This observation is based on the fact that part identities remain constant across multiple views (e.g., the seat in view 1 should be segmented as the seat in view 2, see Fig.~\ref{fig:mask_consistency}). Intuitively, mask consistency can be seen as an additional constraint on point consistency, encouraging points within the same 2D mask to remain associated with the same masks in the 3D space

\begin{figure}[t]
  \vspace{-10pt}
\centering
	\includegraphics[width=0.9\linewidth]{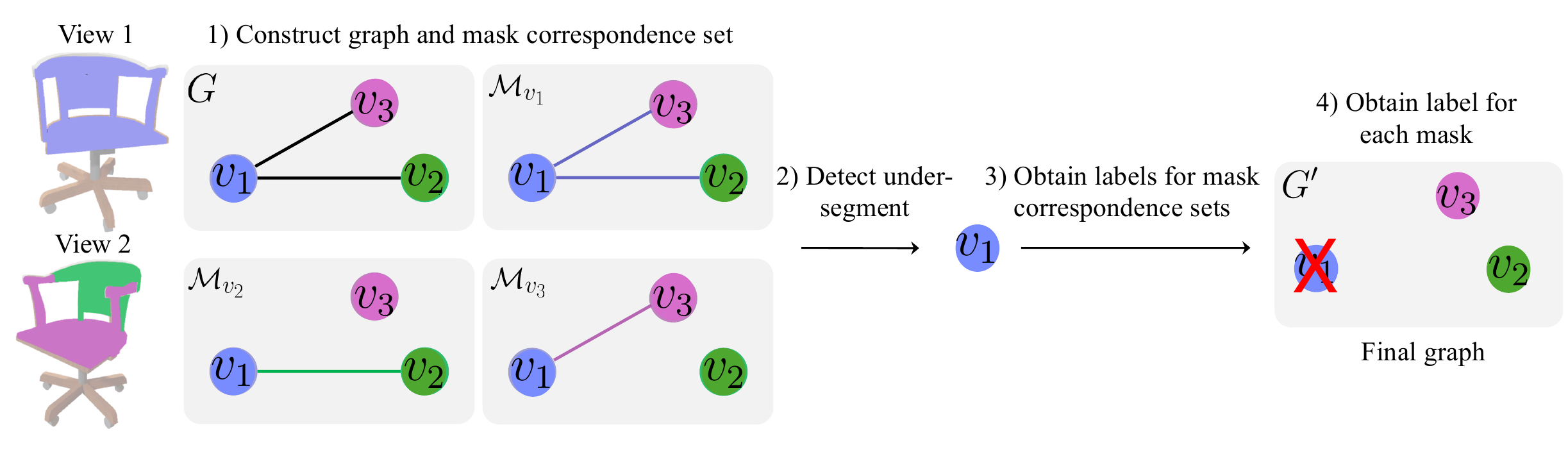}
  \vspace{-10pt}
	\caption{\textbf{Mask-consistency process.} (1) Each vertex of $G$ corresponds to a mask in a given image. The edge connecting each pair of vertices denotes that the pair contains the projection of the same 3D points. Mask consistency set $\mathcal{M}_v$ for each $v$ is obtained via the first-order neighborhood of $v$. (2) $v_1$ is detected as under-segmented since $\mathcal{M}_{v_1}$ consists of masks from the same view with different labels ($v_2$, $v_3$) and hence, is discarded. (3) Traverse $\mathcal{M}_{v_i}$ to obtain labels for $\mathcal{M}_{v_i}$. (4) Obtain label for each mask by majority voting. Here we show a simple example for visualization purpose.}
	\label{fig:mask_consistency_graph}
 \vspace{-20pt}
 
\end{figure}

We present a novel mask-consistency aggregation module that takes a set of 2D part segmentation predictions for multiple views as input. Our approach involves constructing an undirected unweighted graph, denoted as $G:V\rightarrow E$, where each vertex corresponds to a 2D mask in a given view. The edges of the graph connect masks from different views that capture the projection of the same 3D points. We construct a set of mask correspondences for each vertex $v\in V$, $\mathcal{M}_v=\{v,u_1,u_2,\dots u_{N}\}$ where an edge $e_i$ connects $v$ and $u_i$. 
A mask $v$ is defined as undersegmented when there exists at least 2 masks in $\mathcal{M}_v$ that belong to the same image but are assigned with different labels. For instance, in Fig.~\ref{fig:mask_consistency_graph}, vertex $v_1$ corresponds to an undersegmented mask. Formally,
\begin{equation}
\mathcal{S}_v=\{u_i, u_j\in I_k \text{ and } l(u_i)\neq l(u_j) | u_i, u_j\in\mathcal{M}_v\}
\end{equation} where $l(x)$ denotes the label of $x$. We discard $v$ if $|\mathcal{S}_v|>\epsilon$. That is, if $v$ is consistently determined as undersegmented across multiple views, we discard the contribution of $v$ in the final label assignment. We then traverse the graph simultaneously from each vertex using breadth-first-search to accumulate the labels for each $\mathcal{M}_v$. Subsequently, we perform majority voting to assign labels to each $\mathcal{M}_v$. 
Finally, for each mask, we identify the most frequently assigned label as the final label.

The simple intuition behind this approach is: if a part occasionally receives incorrect labels in some challenging views, employing majority voting within the mask correspondence set can calibrate these errors. Further, performing this on the mask level ensures that if two 2D points share the same mask label in the majority of the views, they will ultimately be assigned with the same final label. This approach calibrates potential discrepancies in individual point-wise aggregations (see Fig.~\ref{fig:mc_viz}).



\section{Experiments}\label{sec:experiments}

In this section, we first report the performance of 3-By-2 against baselines on PartNet-Ensembled (PartNetE)~\cite{liu2023partslip} in Sec.~\ref{sec:partnete} and on PartNet~\cite{mo2019partnet} with ``level-3" annotation in Sec.~\ref{sec:partnet}. Note the distinction between these datasets since PartNetE consists of a distinct set of articulated objects from~\cite{xiang2020sapien}. These datasets also exhibit different granularity of part annotations. While PartNetE consists of both basic parts like chair back and fine-grained parts like scissors screw, PartNet with ``level-3" annotation contains all fine-grained parts such as ``back\_frame\_vertical\_bar". In Sec.~\ref{sec:ablation}, we conduct comprehensive ablation studies to verify the necessity of each components in 3-By-2.
Our few-shot experiments refer to the setting where a few labeled 3D objects are available for each object category while there is no annotated 3D part labels in the zero-shot setting. In this setting, we leverage labels from the 2D domain instead.

\begin{table*}[t!]
\begin{center}

\begingroup
\setlength{\tabcolsep}{4pt} 
\renewcommand{\arraystretch}{1.2} 
\caption{\textbf{Few-shot performance on PartNetE~\cite{liu2023partslip} dataset.} The left columns show performance on the 17 categories that supervised methods~\cite{qian2022pointnext,qi2017pointnet++,vu2022softgroup} (first 3 rows) were trained on with additional 28K objects.  The right columns show performance on the 28 categories with only 8 objects/category in the training set.~\cite{singh2018hierarchical,zhao2021few,liu2023partslip,zhou2023partslip++} and ours (last 5 rows) only have access to 8 objects/category during training for all 45 categories. Please refer to the Supplement for the full table on all 45 categories.}
\vspace{-8pt}
\scalebox{0.63}{

\begin{tabular}{l|cccccc|c|cccccc|c|c}
\hline
Methods  & Chair & \begin{tabular}{@{}c@{}}Sci- \vspace{-3pt}\\ ssors \end{tabular} & \begin{tabular}{@{}c@{}}Lap- \vspace{-3pt}\\ top \end{tabular} & Door  & \begin{tabular}{@{}c@{}}Micro- \vspace{-3pt}\\ wave \end{tabular} & \begin{tabular}{@{}c@{}}Key- \vspace{-3pt}\\ board \end{tabular} & \begin{tabular}{@{}c@{}}Avg. \vspace{-3pt}\\ (17) \end{tabular} & \begin{tabular}{@{}c@{}}Cam- \vspace{-3pt}\\ era \end{tabular} & USB  & \begin{tabular}{@{}c@{}}Stap- \vspace{-3pt}\\ ler \end{tabular} & \begin{tabular}{@{}c@{}}Disp- \vspace{-3pt}\\ enser \end{tabular} & \begin{tabular}{@{}c@{}}Ket- \vspace{-3pt}\\ tle \end{tabular} & \begin{tabular}{@{}c@{}}Eye- \vspace{-3pt}\\ gl. \end{tabular} & \begin{tabular}{@{}c@{}}Avg. \vspace{-3pt}\\ (28) \end{tabular} & \begin{tabular}{@{}c@{}}Avg. \vspace{-3pt}\\ (45) \end{tabular}\\ \hline
PointNext~\cite{qian2022pointnext}  & \cellcolor{columbiablue}{0.918} & 0.573 & 0.325 &0.438 & 0.405 & 0.450 & \cellcolor{bisque}{0.591} & 0.332 & \cellcolor{bisque}{0.679} &  \cellcolor{bisque}{0.886} & 0.260 & 0.451 & 0.881 & 0.457 & 0.502\\ 
PointNet++~\cite{qi2017pointnet++}   & 0.847 & 0.500 & \cellcolor{columbiablue}{0.554} & 0.457 &  \cellcolor{bisque}{0.436} & \cellcolor{bisque}{0.745} & 0.533 & 0.065 & 0.524 &  0.516   & 0.121 & 0.209 &  0.762 & 0.250 & 0.365 \\
SoftGroup~\cite{vu2022softgroup}   & \cellcolor{bisque}{0.883} & \cellcolor{columbiablue}{0.760} & 0.184 & \cellcolor{bisque}{0.531} & 0.383 & 0.589 & 0.505 & 0.236 & 0.441 &   0.801  & 0.189 & 0.574 & 0.724 & 0.313 & 0.384\\ \hline
ACD~\cite{singh2018hierarchical}   & 0.390 & 0.391 & 0.111 &  0.189 & 0.066 & 0.261& 0.196 & 0.101 & 0.252  & 0.500 &  0.194  & 0.402 & 0.782  & 0.259 & 0.235\\
Prototype~\cite{zhao2021few}   & 0.708 & 0.430 & 0.279 & 0.334 & 0.270 & 0.449 & 0.419 & 0.320 & 0.654 &  0.807  & 0.534 & 0.607 & 0.779  & 0.470 & 0.451\\
PartSLIP~\cite{liu2023partslip}  & {0.854} & 0.603 & 0.297 &  0.408 & {0.427} & 0.536 & 0.567 & 0.583 & 0.561  & 0.848 & \cellcolor{bisque}{0.738} & 0.770 & \cellcolor{bisque}{0.883}& 0.625 & 0.603\\
PartSLIP++~\cite{zhou2023partslip++}  & 0.853 & 0.605 & 0.297 &  0.451 & \cellcolor{columbiablue}{0.495} & 0.724 & 0.574 & \cellcolor{columbiablue}{0.632} & 0.575   & 0.630  & 0.720 & \cellcolor{columbiablue}{0.856} & \cellcolor{bisque}{0.883} & \cellcolor{bisque}{0.642} & \cellcolor{bisque}{0.615}\\
3-By-2 (ours)  & 0.844 & \cellcolor{bisque}{0.657} & \cellcolor{bisque}{0.453} &  \cellcolor{columbiablue}{0.544} & 0.402 & \cellcolor{columbiablue}{0.896}& \cellcolor{columbiablue}{0.604} & \cellcolor{bisque}{0.626} & \cellcolor{columbiablue}{0.790}  & \cellcolor{columbiablue}{0.901} & \cellcolor{columbiablue}{0.782} & \cellcolor{bisque}{0.815} & \cellcolor{columbiablue}{0.928}& \cellcolor{columbiablue}{0.665} & \cellcolor{columbiablue}{0.642}
\end{tabular}
\label{tab:partnete_few}
}
\endgroup

\vspace{-20pt}
\end{center}

\end{table*}

\subsection{Performance on PartNet-Ensembled}\label{sec:partnete}

\noindent\textbf{Data \& Metric}. We use the dataset provided by Liu et al.~\cite{liu2023partslip} for both the few-shot and zero-shot settings. For each object in both few-shot and test sets, we render 20 RGB images from different views with resolution $800\times 800$. We report mean IoU (mIoU) performance of all baselines using the evaluation protocol provided by~\cite{liu2023partslip} on the input point clouds. Specifically, the performance of a part is not considered if it does not exist in the queried object.


\noindent\textbf{Few-shot Baselines}. We compare 3-By-2 against fully-supervised semantic segmentation~\cite{qian2022pointnext,qi2017pointnet++,vu2022softgroup}, few-shot semantic segmentation~\cite{singh2018hierarchical,zhao2021few} and language-based~\cite{liu2023partslip,zhou2023partslip++} methods. The fully supervised methods~\cite{qian2022pointnext,qi2017pointnet++,vu2022softgroup} were trained on 28K objects of 17 overlapping categories between PartNetE~\cite{liu2023partslip}, in addition to the few-shot set consisting of 8 objects/category. The second group of baselines~\cite{singh2018hierarchical,liu2023partslip,zhou2023partslip++,zhao2021few} were only trained on the few-shot set. PartSLIP and PartSLIP++, a concurrent work, rely on large vision-language model (GLIP~\cite{li2022grounded}) to guide the 2D part detection before extending to the 3D point cloud segmentation. We provide more detailed descriptions in the Supplement. We omit the evaluation of MvDeCor~\cite{sharma2022mvdecor} on this benchmark since it requires ground-truth 3D meshes, whereas PartNetE only provides dense point clouds as inputs.

\noindent\textbf{Few-shot Setting}. In this setting, 8 objects/category serve as the few-shot set. We evaluate on the entire test set of PartNetE~\cite{liu2023partslip}. For a fair comparison, we remove part labels in the test set that do not exist in the few-shot set. We present our few-shot results in Table~\ref{tab:partnete_few}. Compared to fully-supervised 3D methods, we outperform by 1-10\% mIoU on these categories. Additionally, we demonstrate a significant performance boost on the remaining 28 categories (21-41\% mIoU). We further outperform PartSLIP and PartSLIP++ on both subsets, achieving $\sim$3\% mIoU improvements overall.

\noindent\textit{\underline{Performance on Real-world Scans.}} Please note that there is currently no publicly available real-world 3D part segmentation dataset for direct comparison. However, we demonstrate the robustness of our method using real-world objects, as shown in Fig.~\ref{fig:teaser}. These objects were originally introduced by Liu et al.~\cite{liu2023partslip} and captured using an iPhone12 camera.




\noindent\textbf{Zero-shot Baselines}. We compare 3-By-2 with PartSLIP~\cite{liu2023partslip}, VLPart~\cite{sun2023going}-MC and SAMPro3D~\cite{xu2023sampro3d}+OpenMask3D~\cite{takmaz2023openmask3d}. For PartSLIP, we prompt the pre-trained GLIP model with the language inputs without finetuning, following Liu et al.~\cite{liu2023partslip}. VLPart~\cite{sun2023going} is a SOTA 2D part segmentation method that was trained on a combination of various large-scale 2D part datasets. We replace our 2D part segmentation module with a pre-trained VLPart model, retaining the 3D mask-consistency aggregation module as 3-By-2, and term this baseline VLPart-MC. During inference, to guide VLPart effectively, we prompt the model with language inputs as in PartSLIP. SAMPro3D~\cite{xu2023sampro3d} is a SOTA zero-shot instance segmentation method for 3D scenes using SAM at its core. For semantic segmentation evaluation, we integrate SAMPro3D with OpenMask3D~\cite{takmaz2023openmask3d}, an open-vocabulary 3D scene segmentation method.

\noindent\textbf{Zero-shot Setting}. Since we do not have access to any labeled 3D objects in this setting, to effectively transfer part labels, we leverage PACO~\cite{ramanathan2023paco}. This dataset is a fine-grained and richly annotated 2D datasets consisting of objects from COCO-LVIS~\cite{gupta2019lvis}. We crop and mask each annotated object using the provided object bounding box and segmentation mask to form the database. Further, we filter out small objects or objects with limited visibility, using the area of the object segmentation mask as a criterion. 
\begin{table}[t!]
\begin{center}

\begingroup
\setlength{\tabcolsep}{3pt} 
\renewcommand{\arraystretch}{1.2} 
\caption{\textbf{Zero-shot performance} on the subset of PartNetE~\cite{liu2023partslip} that overlaps with PACO~\cite{ramanathan2023paco}. Our method effectively leverages 2D in-the-wild part segmentation dataset to perform 3D part segmentation.}
\vspace{-10pt}
\scalebox{0.7}{

\begin{tabular}{l|cccccccccc|c}
\hline
Methods & \begin{tabular}{@{}c@{}}Ket- \vspace{-3pt}\\ tle \end{tabular}  & \begin{tabular}{@{}c@{}}Micro- \vspace{-3pt}\\ wave \end{tabular}  & \begin{tabular}{@{}c@{}}Sci- \vspace{-3pt}\\ ssors \end{tabular} & \begin{tabular}{@{}c@{}}F.- \vspace{-3pt}\\ Chair \end{tabular} & Mouse & \begin{tabular}{@{}c@{}}Bot- \vspace{-3pt}\\ tle \end{tabular} & Laptop & Clock & Remote & Lamp & \begin{tabular}{@{}c@{}}Avg.- \vspace{-3pt}\\ (18) \end{tabular}\\ \hline
\begin{tabular}{@{}c@{}}SAMPro3D~\cite{xu2023sampro3d}+ \vspace{-3pt}\\ OpenMask3D~\cite{takmaz2023openmask3d} \end{tabular} & 0.026 & 0.001 & 0.118 & 0.437 & 0.019 & 0.103 & 0.017 & 0.007 & 0.084 & 0.074 & 0.146 \\\hline
PartSLIP~\cite{liu2023partslip} & 0.208 & 0.166 & \cellcolor{bisque}0.218 & \cellcolor{columbiablue}{0.917} & \cellcolor{bisque}0.270 &  \cellcolor{bisque} 0.763 & \cellcolor{bisque}0.270 & \cellcolor{columbiablue}{0.267} & 0.115 & \cellcolor{bisque}0.371 & \cellcolor{bisque}0.341\\
VLPart~\cite{sun2023going}-MC & \cellcolor{bisque}0.211 & \cellcolor{bisque}0.192 & 0.193 & \cellcolor{bisque}0.813 & 0.000 & 0.216 & 0.060 & 0.205 & \cellcolor{bisque}0.132 & 0.166 & 0.222\\
3-By-2 (ours)  & \cellcolor{columbiablue}{0.765} & \cellcolor{columbiablue}{0.348} & \cellcolor{columbiablue}{0.594} & 0.712 & \cellcolor{columbiablue}{0.307} & \cellcolor{columbiablue}{0.807} & \cellcolor{columbiablue}{0.394} & \cellcolor{bisque}0.253 & \cellcolor{columbiablue}{0.239} & \cellcolor{columbiablue}{0.500} & \cellcolor{columbiablue}{0.430}\\
\end{tabular}
\label{tab:partnete_zs}
}
\endgroup

\vspace{-20pt}
\end{center}

\end{table}

In Table~\ref{tab:partnete_zs} we show the the performance of all baselines and 3-By-2 on the subset of PartNetE that overlaps with PACO dataset~\cite{ramanathan2023paco}. By leveraging the abundance and fine-grained of 2D in-the-wild part segmentation datasets, we achieve superior performance compared to all baselines (9-29$\%$ mIoU). We significantly outperform PartSLIP on challenging categories with small or thin parts (e.g. scissors and lamp by $28\%$ and $13\%$ mIoU respectively). These results highlight the effectiveness of 3-By-2 even when the database includes challenging real-world images with partial occlusion and truncation.

\begin{table}[t!]
\begin{center}

\begingroup
\setlength{\tabcolsep}{3pt} 
\renewcommand{\arraystretch}{1.2} 
\caption{\textbf{Performance on PartNet dataset} with ``level-3" annotations in the few-shot setting. Bold and underline denote best and second best performance respectively.}
\vspace{-10pt}
\scalebox{0.7}{

\begin{tabular}{l|cccccccccc|c}
\hline
Methods & \begin{tabular}{@{}c@{}}Bot- \vspace{-3pt}\\ tle \end{tabular}  & \begin{tabular}{@{}c@{}}Micro- \vspace{-3pt}\\ wave \end{tabular}  & \begin{tabular}{@{}c@{}}Dis- \vspace{-3pt}\\ play \end{tabular} & \begin{tabular}{@{}c@{}}Dish- \vspace{-3pt}\\ washer \end{tabular} & \begin{tabular}{@{}c@{}}Fau- \vspace{-3pt}\\ cet \end{tabular} & Knife & \begin{tabular}{@{}c@{}}Ear- \vspace{-3pt}\\ phone \end{tabular} & Clock & Bed & \begin{tabular}{@{}c@{}}Trash- \vspace{-3pt}\\ can \end{tabular} & Avg.\\ \hline
MvDeCor~\cite{sharma2022mvdecor} & \cellcolor{bisque}{0.421} & \cellcolor{bisque}{0.377} & \cellcolor{columbiablue}{0.600} & 0.327 & \cellcolor{columbiablue}{0.212} & \cellcolor{bisque}{0.187} & \cellcolor{bisque}{0.205} & \cellcolor{columbiablue}{0.143} & \cellcolor{columbiablue}{0.099} & \cellcolor{columbiablue}{0.199} & \cellcolor{bisque}{0.277}\\ 
PartSLIP~\cite{liu2023partslip} & 0.344 & 0.143 & 0.386 & 0.228 & 0.009 & 0.023 & 0.064 & 0.017 & 0.003 & 0.031 & 0.125\\
3-By-2 (ours)  & \cellcolor{columbiablue}{0.454} & \cellcolor{columbiablue}{0.389} & \cellcolor{bisque}{0.567} & \cellcolor{columbiablue}{0.429} &  \cellcolor{bisque}{0.203} & \cellcolor{columbiablue}{0.196} & \cellcolor{columbiablue}{0.225} & \cellcolor{bisque}{0.116} & \cellcolor{bisque}{0.096} & \cellcolor{bisque}{0.134} & \cellcolor{columbiablue}{0.281}\\
\end{tabular}
\label{tab:partnet}
}
\endgroup

\vspace{-25pt}
\end{center}

\end{table}

\noindent\textit{\underline{Effectiveness of Our 2D Segmentation Module.}} We demonstrate the effectiveness of our 2D segmenter, leveraging SAM and DIFT, by showcasing its strong performance against VLPart~\cite{sun2023going}, a SOTA 2D part segmentation method (see Table~\ref{tab:partnete_zs}, last 2 rows). Note that VLPart was trained on PACO~\cite{ramanathan2023paco} among other 2D part datasets. Therefore, it is reasonable to anticipate that this method can effectively use knowledge from PACO to accurately segment the 18 overlapping categories between PartNetE and PACO. For both VLPart-MC and 3-By-2, we maintain the same 3D aggregation module. Our method significantly outperforms VLPart-MC, demonstrating the advantage of our proposed 2D segmentation module.

\noindent\textit{\underline{Comparison to SOTA Scene Segmentation Approach.}} SAMPro3D~\cite{xu2023sampro3d} is a concurrent work with SOTA performance on zero-shot instance segmentation in 3D scene. This is a training-free model that effectively prompts SAM within the 2D domain using 3D point projections. As in Table~\ref{tab:partnete_zs}, we outperform this baseline by a significant margin, highlighting the non-trivial nature of adapting scene segmentation methods for 3D part segmentation tasks, particularly those involving post-processing of 2D foundation models.

\subsection{Performance on Level-3 PartNet}\label{sec:partnet}
In this experiment, we select 10 categories from PartNet~\cite{mo2019partnet} that come with fine-grained (``level-3") annotations. We randomly select 10 objects per category from the training set (following~\cite{sharma2022mvdecor}) to form our few-shot set, and up to 50 objects per category from the test set for evaluation, ensuring overlap with ShapeNetCore.v2~\cite{chang2015shapenet}. Given that PartSLIP~\cite{liu2023partslip} employs point cloud RGB for superpoint generation, which serves as 3D priors, our decision to choose overlapping objects with ShapeNetCore.v2 is to preserve object texture information. We use the same few-shot and test set for all baselines. 

\noindent\textbf{Data}. As inputs to our approach, we render 15 overlapping views for each textured mesh using Blender cycle renderer with realistic lighting from HDRI environment maps.

\noindent\textbf{Baselines}. The baselines are reproduced following the papers' recommended training procedure. Specifically, we pre-train MvDeCor~\cite{sharma2022mvdecor} on the entire training set of the selected categories consisting of 86 views per non-textured object, with rendered RGB, depth and normal maps as inputs. We then fine-tune the segmentation heads for each individual object category in the few-shot set with 15 views per object. Note that the input for this stage also includes RGB, depth and normal maps.

For PartSLIP~\cite{liu2023partslip}, we derive the language prompt by traversing the part hierarchy and concatenating labels from each level along the path, spanning from root to leaf. For example, the path "bottle/jug/handle" is transformed into "bottle jug handle". This adaptation is due to the potential for different leaf nodes to share identical labels (e.g., bottle/normal\_bottle/handle and bottle/jug/handle), as relying solely on the leaf node label could introduce confusion in predictions. We adopt PartSLIP's point cloud, image rendering and data processing pipeline with default parameters.

\noindent\textbf{Evaluation \& Metric}. We uniformly sample 300K points on the surface of each labeled ground truth mesh and employ nearest neighbor assignment to associate a ground-truth label with each point. This point set is used for evaluating all methods for a fair comparison and eliminating any randomness introduced by the point cloud sampling step. We use part mIoU on the sampled point set as the evaluation metric. We employ the standard mIoU calculation, which considers the performance of all parts in the vocabulary, even in cases where they may not exist in certain objects. Additionally, different from MvDeCor, we do not exclude the ``others" label during evaluation based on ground-truth labels. For a fair comparison, we applied the same evaluation approach across all methods.

\noindent\textbf{Results.} We show results in Table~\ref{tab:partnet}. Compared to PartSLIP~\cite{liu2023partslip}, we outperformed on all categories by a significant margin (16\% mIoU on average), demonstrating the challenges posed by fine-grained settings for GLIP~\cite{li2022grounded}. While our performance is on par with MvDeCor~\cite{sharma2022mvdecor}, it is important to note that MvDeCor is both pretrained and finetuned on PartNet~\cite{mo2019partnet}, using ground truth depth and normal maps as additional inputs. In contrast, our method requires no training on the target data distribution.


\subsection{Ablation Study}\label{sec:ablation}
\noindent\textbf{Non-overlapping Mask Generation.} In Table~\ref{tab:non_overlap_ablation}, we illustrate the effectiveness of our proposed non-overlapping mask generation module. The comparison involves evaluating the performance of our method with and without this module. In the case of the model without the non-overlapping mask generation module, we directly utilize the predicted SAM outputs for label transferring. The results indicate that our non-overlapping mask generation module is necessary for achieving an optimal performance.

\begin{table}[t]
\begin{minipage}{0.51\textwidth}
\centering

\begin{center}

\begingroup
\setlength{\tabcolsep}{3pt} 
\renewcommand{\arraystretch}{1.2} 
\caption{Ablation of the non-overlapping mask generation module.}
\vspace{-8pt}
\scalebox{0.68}{

\begin{tabular}{l|ccccc}
\hline
2D Mask Proposal & Scissors & Mouse & Suitcase & Bottle & Chair\\
\hline
 SAM & 0.457 & 0.440 & 0.285 & 0.004 & 0.638 \\
\hline
Non-overlap & \cellcolor{columbiablue}{0.675} &  \cellcolor{columbiablue}{0.684} & \cellcolor{columbiablue}{0.813} & \cellcolor{columbiablue}{0.810} & \cellcolor{columbiablue}{0.844}\\
\end{tabular}

\label{tab:non_overlap_ablation}
}
\endgroup

\end{center}


    \label{table:task}
\end{minipage}
\hspace{4pt}
\begin{minipage}{0.49\textwidth}
\centering
\begin{center}

\begingroup
\setlength{\tabcolsep}{3pt} 
\renewcommand{\arraystretch}{1.2} 
\caption{Ablation of our proposed mask-consistency component.}
\vspace{-18pt}
\scalebox{0.7}{

\begin{tabular}{l|cccc}
\hline
3D Label Aggregation & Scissors & Suitcase & Printer & Clock\\
\hline
 Point-Consistency & 0.619 & 0.579 & 0.009 & 0.363 \\
\hline
Mask-Consistency & \cellcolor{columbiablue}{0.675} &  \cellcolor{columbiablue}{0.684} & \cellcolor{columbiablue}{0.085} & \cellcolor{columbiablue}{0.458}\\
\end{tabular}

\label{tab:mc_sup}
}
\endgroup

\end{center}


    \label{table:task}
\end{minipage}
\vspace{-20pt}
\end{table}
\noindent\textbf{Mask-consistency Module.} In Table~\ref{tab:mc_sup}, we demonstrate the effectiveness of our proposed mask-consistency component, which improves the final performance especially on objects with small parts.

\noindent\textbf{Properties of Database.} In this section, we investigate two key questions: 1) Can 3-By-2 accurately segment the query object within a database containing multiple object categories? and 2) Is it possible to transfer parts with the same semantic meaning between different object categories?

\noindent\textit{\underline{Multi-category database.}} To address question 1, we perform experiments using databases containing 1, 2, and 8 categories respectively (see Table~\ref{tab:multi}). 
Specifically, taking the query category as ``Kettle", for the 2-category setting we construct a database consisting of "Kettle, Kitchen Pot". We selected these categories due to their shared semantic parts with "Kettle", which could potentially lead to confusion (e.g., kettle lid vs. kitchen pot lid). With 8-category setting, we add in categories that are completely different and do not share any parts with ``Kettle" (e.g. ``Eyeglasses"). In general, with more categories in the database, there is a slight decrease in the average performance. Notably, there are marginal differences between 2-category and 8-category (second and third rows), highlighting the the ability of 3-By-2 in handling both diverse object taxonomy and part segmentation. This finding is particularly interesting since many prior works~\cite{liu2023partslip,sharma2022mvdecor} require finetuning each category separately for few-shot evaluation.

\noindent\textit{\underline{Cross-category database.}} Considering question 2, we note that the few-shot set of ``Table" in PartNetE lacks objects with wheels as a part, whereas such objects are present in the test set. To address this, we incorporate the ``Chair" category where the wheel part exists in the database. We evaluate on 18 tables in PartNetE test set with the ``wheel" part annotated (see Table~\ref{tab:cross}). 
Compared to the table only few-shot set, combining the database with ``Chair" improves the performance on ``leg" by $\sim6\%$ mIoU. The improvement in the ``leg" part can be attributed to the inclusion of ``Chair" in the database, which reduces the likelihood of the model incorrectly associating ``wheel" with ``leg" due to the absence of ``wheel" in the few-shot set. Interestingly, the performance for ``wheel" increases significantly, $+60\%$ mIoU through the label transfer from chair wheels.  


\begin{table}[t]
\begin{minipage}{0.55\textwidth}

\begin{center}

\begingroup
\setlength{\tabcolsep}{3pt} 
\renewcommand{\arraystretch}{1.2} 
\caption{\textbf{Multi-category database experiment.} Performance of Kettle in various database settings is reported with mIoU. Our method shows robustness in performance even when more categories are added in the database.}
\scalebox{0.8}{

\begin{tabular}{l|cccc}
\hline

Database & Lid & Handle & Spout & Avg.\\
\hline
1-category & 0.759 & 0.904 & 0.783 & 0.815 \\
2-category & 0.703 & 0.820 & 0.748 & 0.757 \\
8-category & 0.727 & 0.773 & 0.756 & 0.752
\end{tabular}
\label{tab:multi}
}
\endgroup

\vspace{-20pt}
\end{center}

\end{minipage}
\hspace{5pt}
\begin{minipage}{0.46\textwidth}

\begin{center}

\begingroup
\setlength{\tabcolsep}{3pt} 
\renewcommand{\arraystretch}{1.2} 
\caption{\textbf{Cross-category database experiment.} We report the performance of 18 tables with wheels in PartNetE. Results show that our method can transfer wheel annotations from Chair to correct the prediction on Table wheels.}
\vspace{-12pt}
\scalebox{0.7}{

\begin{tabular}{l|ccc|c}
\hline
Database  & Leg & Tabletop & Wheel & Avg. \\
\hline
Table only & 0.586 & 0.647 & 0.000 & 0.411\\
Chair \& Table & \cellcolor{columbiablue}0.641 & 0.633 & \cellcolor{columbiablue}0.600 & \cellcolor{columbiablue}{0.625}
 
\end{tabular}
\label{tab:cross}
}
\endgroup

\vspace{-15pt}
\end{center}


\end{minipage}
\end{table}

While the concept may seem intuitive, our findings shed new light on object part compositionality. Despite the diversity in appearances and shapes across various object categories, there exists a finite set of object parts that are shared among them. Recognizing the transferability of these parts is important for facilitating rapid learning of novel objects across a range of tasks. Further, our results show the ability to correct wrong predictions of our approach by transferring labels from another category. Please refer to the Sup. for additional studies.


\section{Conclusion}
In this work, we propose \textbf{3-By-2}, a novel, training-free method that achieves SOTA performance on benchmarks with diverse levels of part granularity without the need for language inputs, on both zero-shot and few-shot settings. We demonstrate the flexibility of 3-By-2 in transferring part labels between different object categories. We hope the development of 3-By-2 can encourage further exploration of visual similarities for this task.
\clearpage
\section*{Acknowledgement} This work was partly supported by NIH R01HD104624-01A1.
\clearpage
This Supplementary Material is structured as follows: in Section~\ref{sec_sup:data} we provide more details about the datasets, the data rendering pipelines and ground-truth processing; in Section~\ref{sec_sup:results} we present full results for the PartNetE few-shot and zero-shot experiments in the main text; in Section~\ref{sec_sup:ablation} we perform additional ablation studies for our method; in Section~\ref{sec_sup:implementation}, we provide implementation details of our approach; in Section~\ref{sec_sup:viz}, we demonstrate the effectiveness of the semantic correspondence matching process; in Section~\ref{sec_sup:discussion} we discuss about the limitation of our work and potential future directions; in Section~\ref{sec_sup:qualitative}, we present additional qualitative results.

\section{Data}\label{sec_sup:data}
\subsection{Datasets}
\textbf{PartNet-Ensembled~\cite{liu2023partslip}.} This dataset combines objects from the PartNet-Mobility dataset~\cite{xiang2020sapien} for few-shot and test splits, along with additional training objects from PartNet~\cite{mo2019partnet}. In our study, we exclusively use the few-shot and test splits. The few-shot split encompasses 8 objects per category across all 45 categories, while the test split spans from 6 to 338 objects per category, having 1906 objects in total. PartNetE consists of 106 parts, ranging from basic components like chair backs to more fine-grained parts such as scissors screws. Note that compared to the level-3 annotations of PartNet~\cite{mo2019partnet}, PartNetE exhibits a coarser granularity and consists of 28 additional categories from PartNet Mobility dataset~\cite{xiang2020sapien}. 

\begin{figure*}[t]
\centering
	\includegraphics[width=\linewidth]{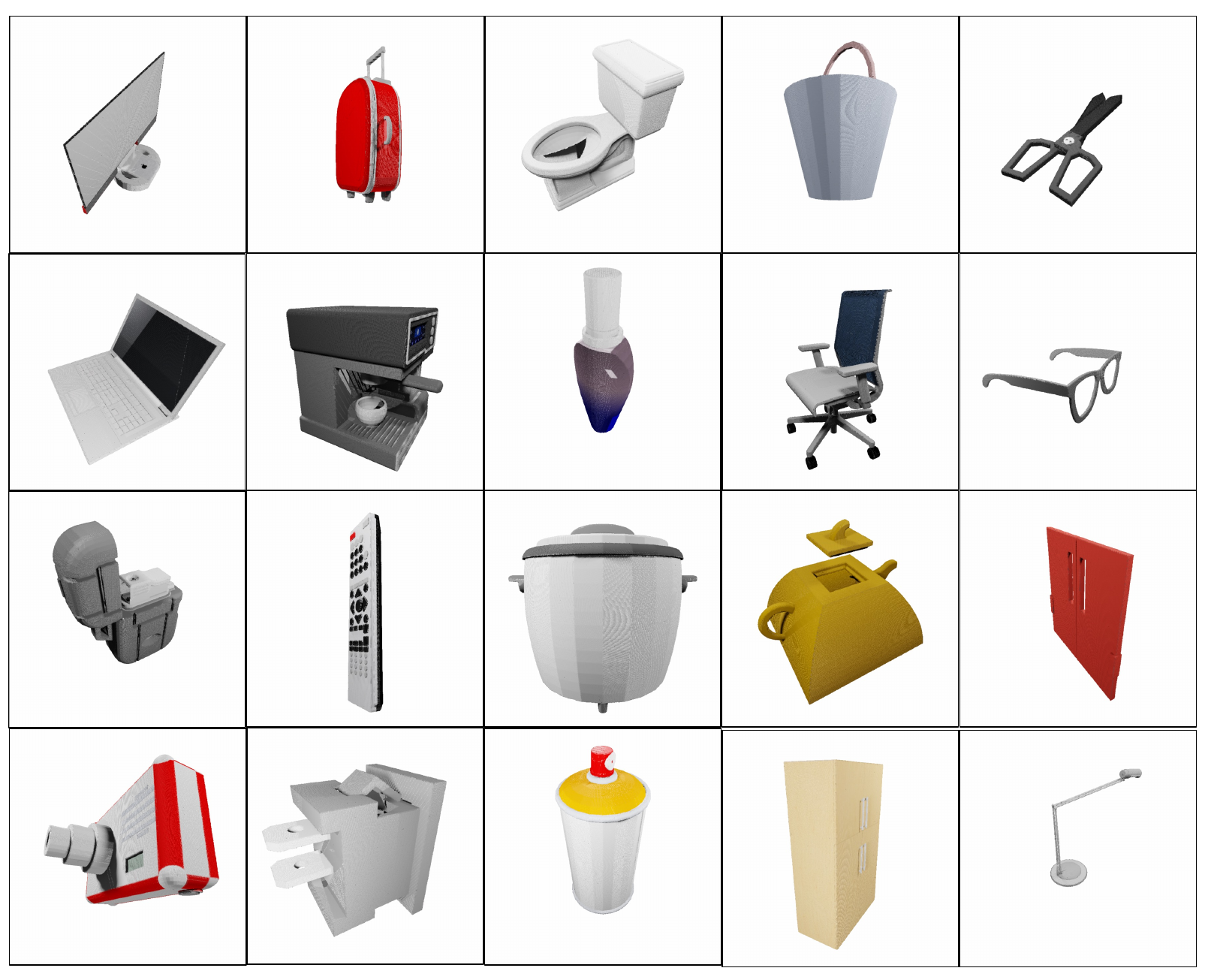}
	\caption{Rendered images for 3-By-2 on PartNetE~\cite{liu2023partslip}. Each object is rendered with resolution $800\times 800$ images on a white background.}
	\label{fig:partnetE_inp_sup}
\end{figure*}

\noindent\textbf{PartNet~\cite{mo2019partnet}.} This dataset consists of annotations at various levels of granularity, spanning from level-1 to level-3, where level-3 represents the most fine-grained part annotations, including labels such as ``back\_frame\_vertical\_bar" for chairs. We perform experiments on 10 object categories with level-3 part annotations. As mentioned in the main text, we use a subset of PartNet that overlaps with ShapeNet~\cite{chang2015shapenet}.

\subsection{Data Rendering}
\textbf{PartNetE Experiments.} For these experiments, we use the rendering code provided by Liu et al.~\cite{liu2023partslip}. Specifically, we employ PyTorch3D point cloud rasterization to render the input point clouds. We render 20 RGB views for each point cloud on a white background. Pixel and 3D point correspondences are saved for inference. Refer to Figure~\ref{fig:partnetE_inp_sup} for examples of rendered images.

\begin{figure*}[t]
\centering
	\includegraphics[width=\linewidth]{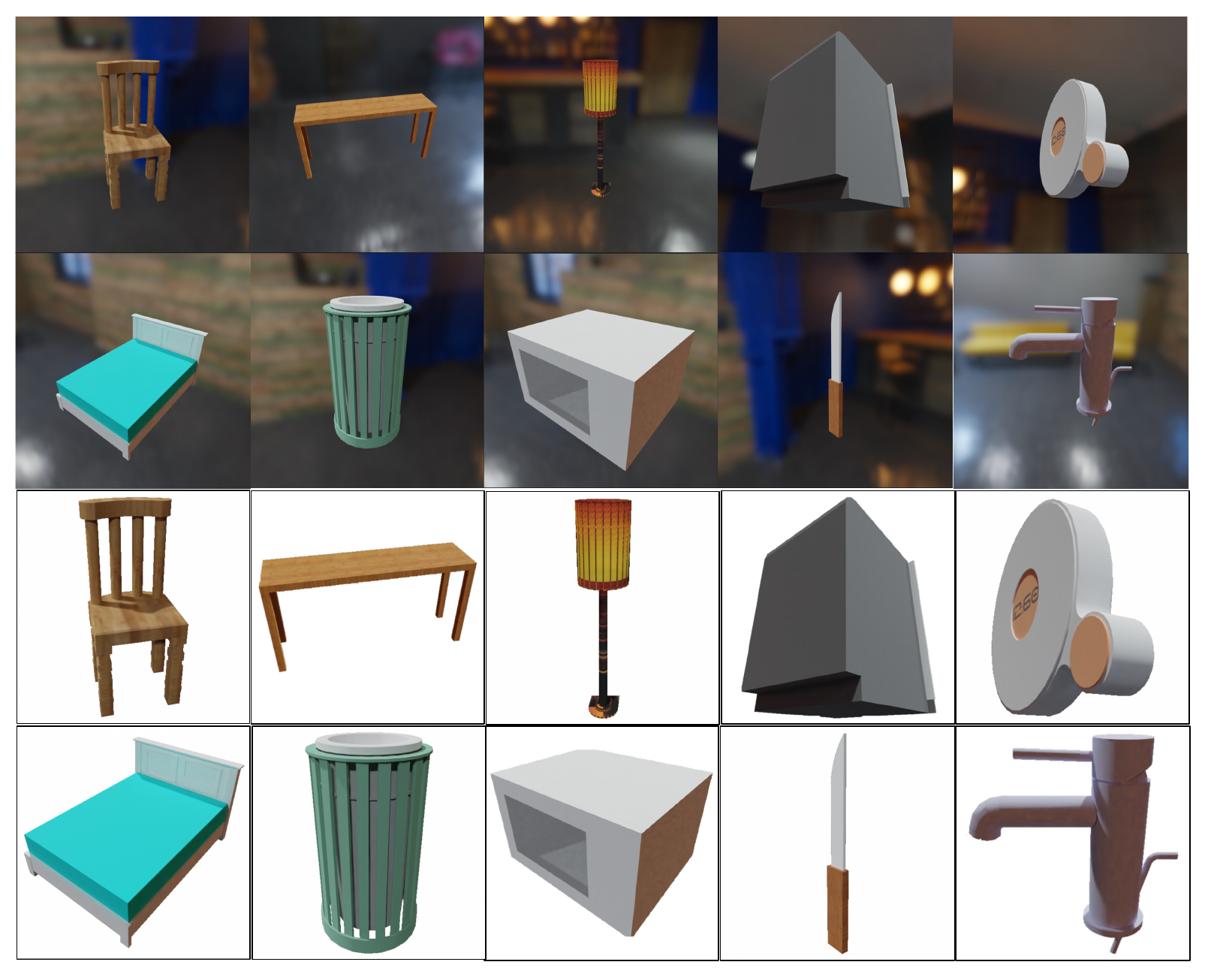}
	\caption{(Top) Rendered images for 3-By-2 on PartNet~\cite{mo2019partnet}. We use Blender to render each object with photorealistic lightings from HDRI environment maps. (Bottom) Square cropped and masked inputs used by 3-By-2 for label transferring.}
	\label{fig:partnet_inp_sup}
\end{figure*}

\noindent\textbf{PartNet Experiment.} For our method, we employ Blender~\cite{blender} to render textured meshes from ShapeNet~\cite{chang2015shapenet}, incorporating an HDRI environment map to ensure photorealistic lighting variations (refer to Figure~\ref{fig:partnet_inp_sup}, 2 top rows). For each object, we generate 15 RGB views with resolution $800\times 800$. At inference, we use the object's ground-truth segmentation mask to mask out the background. We additionally center and square crop the objects based on their bounding boxes, with a padding of 16 pixels on each side (Figure~\ref{fig:partnet_inp_sup}, 2 bottom rows).

For PartSLIP~\cite{liu2023partslip}, we first convert each textured mesh into textured point cloud using BlenderProc~\cite{Denninger2023}. We then use the provided point cloud rendering code from PartSLIP to render input images with default rendering parameters.

For MvDeCor~\cite{sharma2022mvdecor}, we follow the rendering code provided by the authors to render RGB, depth and normal maps for each untextured mesh. For pre-training, we follow the default rendering parameters from~\cite{sharma2022mvdecor} and render 86 views on a white background with resolution $256\times 256$. During low-shot finetuning and inference, we use 15 views per object.

\subsection{Ground-truth Processing}
For PartNetE experiments, we use the ground-truth labels provided by Liu et al.~\cite{liu2023partslip}. In the case of PartNet, we follow~\cite{sharma2022mvdecor}, where we initially convert the ground-truth labels for 1024 points to labels for triangle faces of the processed object mesh. Subsequently, we sample a dense point cloud of 300K points from the surface of each object and transfer the labels from the mesh to this standardized point cloud via querying the nearest face neighbor for each point. This point cloud is used to evaluate the performance of all methods.

\section{Full Table Results}\label{sec_sup:results}
\subsection{Metric Details}
For each object category, we first record the average IoU for each object part. The mean IoU reported for each category is the average IoU of all the parts in that category.

The standard Intersection-over-Union (IoU) metric is computed as the following:
\[
IoU_\text{standard} = \frac{m_\text{pred.}\cap m_\text{gt}}{m_\text{pred.}\cup m_\text{gt}}
\]
where $m_\text{pred.}$ and $m_\text{gt}$ are the binary masks for the predicted and ground-truth segmentation respectively. We employ this implementation of IoU in our PartNet experiment.

In the PartNetE experiments, we follow the evaluation protocol outlined in~\cite{liu2023partslip} to ensure a fair comparison with baselines and maintain consistencies. Specifically, we omit the prediction of a part if that part does not actually exist in the object. Formally,
\[
IoU = \mathds{1}\{m_\text{gt}\neq\bold{0}\}\cdot\frac{m_\text{pred.}\cap m_\text{gt}}{m_\text{pred.}\cup m_\text{gt}}
\]

\subsection{Few-shot on PartNetE} In this setting, we compare our method against 3D fully-supervised semantic segmentation methods: PointNext~\cite{qian2022pointnext}, PointNet++~\cite{qi2017pointnet++} and SoftGroup~\cite{vu2022softgroup}; few-shot semantic segmentation methods: ACD~\cite{gadelha2020label} and Prototype~\cite{zhao2021few}; language-based part segmentation methods: PartSLIP~\cite{liu2023partslip} and PartSLIP++~\cite{zhou2023partslip++}. 

The first group of methods was trained fully-supervised on all objects of 17 categories in PartNet~\cite{mo2019partnet}, in addition to the few-shot set of PartNetE, which consists of 8 objects for each of 45 categories. ACD and Prototype exclusively trained on the few-shot subset of PartNetE. While ACD~\cite{gadelha2020label} integrates a proxy self-supervised task to facilitate few-shot learning, Prototype~\cite{zhao2021few} extracts features serving as prototypes for each part category. In contrast, PartSLIP~\cite{liu2023partslip} and PartSLIP++~\cite{zhou2023partslip++} relies on large vision-language model (GLIP~\cite{li2022grounded}) to guide the 2D part detection before extending to the 3D point cloud segmentation. All these baselines require either fine-tuning or training on the few-shot set, whereas our approach is training-free, using the few-shot set solely for constructing the database to transfer labels. We present the performance on all 45 categories of PartNetE in the few-shot setting in Table~\ref{tab:partnete_few_sup} and Table~\ref{tab:partnete_few_sup2}.


\begin{table*}[t!]
\begin{center}

\begingroup
\setlength{\tabcolsep}{2.5pt} 
\renewcommand{\arraystretch}{1.2} 
\caption{Few-shot performance on PartNetE~\cite{liu2023partslip} dataset on the 17 categories that overlap between PartNet~\cite{mo2019partnet} and PartNetE. The first three columns show the performance of fully-supervised methods. The right columns show performance of the methods that only have access to the few-shot set.}
\scalebox{0.7}{

\begin{tabular}{|l|ccc|ccccc|}
\hline
Categories $\downarrow$ & \begin{tabular}{@{}c@{}}PointNet++ \vspace{-3pt}\\ \cite{qi2017pointnet++} \end{tabular}  & \begin{tabular}{@{}c@{}}SoftGroup \vspace{-3pt}\\ \cite{vu2022softgroup} \end{tabular} & \begin{tabular}{@{}c@{}}PointNext \vspace{-3pt}\\ \cite{qian2022pointnext} \end{tabular} & \begin{tabular}{@{}c@{}}ACD \vspace{-3pt}\\ \cite{gadelha2020label} \end{tabular}  & \begin{tabular}{@{}c@{}}Prototype \vspace{-3pt}\\ \cite{zhao2021few} \end{tabular} & \begin{tabular}{@{}c@{}}PartSLIP \vspace{-3pt}\\ \cite{liu2023partslip} \end{tabular} & \begin{tabular}{@{}c@{}}PartSLIP++ \vspace{-3pt}\\ \cite{zhou2023partslip++} \end{tabular}  & 3-By-2\\
\hline
Bottle & 0.488 & 0.414 & 0.684 & 0.224 & 0.601 & \cellcolor{bisque}0.834 & \cellcolor{columbiablue}0.855 & 0.809\\
Chair & 0.847 & \cellcolor{bisque}0.883 & \cellcolor{columbiablue}{0.918} & 0.390 & 0.708 & 0.854 & 0.853 & 0.844 \\
Clock & 0.192 & 0.025 & 0.284 & 0.000 & 0.105 & 0.376 & \cellcolor{columbiablue}0.541 &\cellcolor{bisque}{0.458} \\
Dishwasher & 0.495 & 0.530 & \cellcolor{columbiablue}{0.692} & 0.253 & 0.483 & \cellcolor{bisque}{0.625} & 0.609 &0.536\\
Display & 0.783 & 0.621 & \cellcolor{columbiablue}{0.894} & 0.291 & 0.673 & \cellcolor{bisque}{0.848} & 0.851 & 0.726 \\
Door & 0.457 & \cellcolor{bisque}0.531 & 0.438 & 0.189 & 0.334 & 0.408 & 0.451 &\cellcolor{columbiablue}{0.544} \\
Faucet & 0.672 & 0.684 & \cellcolor{columbiablue}{0.850} & 0.242 & 0.460& \cellcolor{bisque}{0.714} & 0.659 &0.669 \\
Keyboard & \cellcolor{bisque}0.745 & 0.589 & 0.450 & 0.261 & 0.449 & 0.536 & 0.724 &\cellcolor{columbiablue}{0.896} \\
Knife &0.354 & 0.313 & 0.587 & 0.396 & 0.504 & \cellcolor{bisque}{0.652} & 0.643 & \cellcolor{columbiablue}{0.751} \\
Lamp & \cellcolor{bisque}0.680 & \cellcolor{columbiablue}0.822 & 0.649 & 0.137 & 0.382 & 0.661 & \cellcolor{bisque}0.680 & 0.595 \\
Laptop & \cellcolor{columbiablue}0.554 & 0.184 & 0.325 & 0.111 & 0.279 & 0.297 & 0.297 & \cellcolor{bisque}0.453 \\
Microwave & \cellcolor{bisque}0.436 & 0.383 & 0.405 & 0.066 & 0.270 & 0.427 & \cellcolor{columbiablue}0.495 & 0.402\\
Refrigerator & 0.434 & 0.469 & \cellcolor{columbiablue}{0.762} & 0.108 & 0.429 & \cellcolor{bisque}{0.558} & 0.557 & 0.517\\
Scissors & 0.500 & \cellcolor{columbiablue}0.760 & 0.573 & 0.391 & 0.430 & 0.603 & 0.605 & \cellcolor{bisque}0.657 \\
StorageFurniture & 0.469 & \cellcolor{bisque}0.602 & \cellcolor{columbiablue}{0.685} & 0.076 & 0.302 & 0.536 & 0.573 & 0.517\\
Table & 0.577 & \cellcolor{bisque}0.591 & \cellcolor{columbiablue}{0.616} & 0.202 & 0.386 & 0.484 & 0.521 & 0.566 \\
TrashCan & \cellcolor{columbiablue}0.717 & 0.170 & 0.228 & 0.000 & 0.329 & 0.223 &0.242 & \cellcolor{bisque}0.331 \\
\hline
Avg. (17) & 0.533 & 0.505 & \cellcolor{bisque}{0.591} & 0.196 & 0.419 & 0.567 & 0.574 & \cellcolor{columbiablue}{0.604} \\
\hline
\end{tabular}
\label{tab:partnete_few_sup}


\label{tab:partnete_few_sup}

}
\endgroup

\end{center}

\end{table*}

\begin{table*}[t!]
\begin{center}

\begingroup
\setlength{\tabcolsep}{2.5pt} 
\renewcommand{\arraystretch}{1.2} 
\caption{Few-shot performance on PartNetE~\cite{liu2023partslip} dataset on the remaining 28 categories. The first three columns show the performance of fully-supervised methods. The right columns show performance of the methods that only have access to the few-shot set.}
\scalebox{0.7}{
\begin{tabular}{|l|ccc|ccccc|}
\hline
Categories $\downarrow$ & \begin{tabular}{@{}c@{}}PointNet++ \vspace{-3pt}\\ \cite{qi2017pointnet++} \end{tabular}  & \begin{tabular}{@{}c@{}}SoftGroup \vspace{-3pt}\\ \cite{vu2022softgroup} \end{tabular} & \begin{tabular}{@{}c@{}}PointNext \vspace{-3pt}\\ \cite{qian2022pointnext} \end{tabular} & \begin{tabular}{@{}c@{}}ACD \vspace{-3pt}\\ \cite{gadelha2020label} \end{tabular}  & \begin{tabular}{@{}c@{}}Prototype \vspace{-3pt}\\ \cite{zhao2021few} \end{tabular} & \begin{tabular}{@{}c@{}}PartSLIP \vspace{-3pt}\\ \cite{liu2023partslip} \end{tabular} & \begin{tabular}{@{}c@{}}PartSLIP++ \vspace{-3pt}\\ \cite{zhou2023partslip++} \end{tabular}  & 3-By-2\\
\hline
Box & 0.186 & 0.088 & 0.842 & 0.211 & 0.686 & \cellcolor{bisque}{0.845} & \cellcolor{columbiablue}0.855 & {0.761}\\
Bucket & 0.000 & 0.250 & 0.041 & 0.000 & 0.313 & {0.365} & \cellcolor{columbiablue}0.855 & \cellcolor{bisque}0.784\\
Camera & 0.065 & 0.236 & 0.332 & 0.101 & 0.320 & {0.583} & \cellcolor{columbiablue}0.632 & \cellcolor{bisque}{0.626} \\
Cart & 0.064 & 0.239 & 0.363 & 0.315 & 0.368 & \cellcolor{columbiablue}{0.881} & \cellcolor{bisque}0.849 &{0.812}\\
CoffeeMachine & 0.346 & 0.083 & 0.179 & 0.007 & 0.200 & \cellcolor{columbiablue}{0.378} & \cellcolor{bisque}0.376 & {0.342}\\
Dispenser & 0.121 & 0.189 & 0.260 & 0.194 & 0.534 & \cellcolor{bisque}{0.738} & 0.720 &\cellcolor{columbiablue}{0.782}\\
Eyeglasses & 0.762& 0.724 & 0.881 & 0.782 & 0.779 & \cellcolor{bisque}{0.883} & \cellcolor{bisque}0.883 &\cellcolor{columbiablue}{0.928} \\
FoldingChair & 0.109 & 0.147 & \cellcolor{columbiablue}{0.964} & 0.742& 0.912 & 0.863 & 0.899  &\cellcolor{bisque}{0.936}\\
Globe & 0.465 & 0.590 & 0.923 & 0.698& 0.883& \cellcolor{columbiablue}{0.957} &0.928 & \cellcolor{bisque}{0.952} \\
Kettle & 0.209 & 0.574 & 0.451 & 0.402 & 0.607 & {0.770} & \cellcolor{columbiablue}0.856 &\cellcolor{bisque}{0.815} \\
KitchenPot & 0.158 &0.455  & 0.570 & 0.519 &0.633 & \cellcolor{bisque}{0.696} & \cellcolor{columbiablue}0.729& {0.650} \\
Lighter & 0.350& 0.302& 0.163 & 0.108&0.307 & {0.647} & \cellcolor{columbiablue}0.662 &\cellcolor{bisque}{0.650}\\
Mouse & 0.167& 0.559& 0.325 & 0.000&0.301 & {0.580} & \cellcolor{bisque}0.587& \cellcolor{columbiablue}{0.684}\\
Oven & 0.344& 0.137& 0.378 & 0.089& 0.365& \cellcolor{columbiablue}{0.735} & \cellcolor{bisque}0.703 &{0.600}  \\
Pen &0.465 & 0.289& 0.395 &0.054 & 0.475& \cellcolor{columbiablue}{0.715} & \cellcolor{bisque}0.664& {0.628}\\
Phone &0.200 & 0.024& 0.395 &0.099 & 0.355& \cellcolor{bisque}{0.484} & \cellcolor{columbiablue}0.590 &{0.410} \\
Pliers & 0.577 & 0.742 & \cellcolor{columbiablue}{0.996} & 0.835 & 0.910 & 0.332 & 0.297 & \cellcolor{bisque}{0.994} \\
Printer & 0.000 & 0.012 & 0.000 &0.000 & 0.002 & {0.043} & \cellcolor{bisque}0.062  &\cellcolor{columbiablue}{0.085} \\
Remote & 0.036 & 0.371& \cellcolor{columbiablue}{0.578} & 0.000 & 0.296 & 0.383 & 0.364 &\cellcolor{bisque}{0.541} \\
Safe &0.319 & 0.053& 0.256 & 0.093&0.201 & {0.322} & \cellcolor{columbiablue}0.372 &\cellcolor{bisque}{0.347} \\
Stapler & 0.516 & 0.801 & \cellcolor{bisque}{0.886} & 0.500 & 0.807 & 0.848 & 0.630 &\cellcolor{columbiablue}{0.901} \\
Suitcase &0.407 & 0.183& 0.136 & 0.132&0.355 & \cellcolor{columbiablue}{0.704} & \cellcolor{bisque}0.700 &{0.652} \\
Switch & 0.018& 0.210 & 0.397 &0.103 & 0.409 & \cellcolor{columbiablue}{0.594} & \cellcolor{bisque}0.561 & {0.546} \\
Toaster & 0.147& 0.264& 0.084 & 0.000& 0.101& \cellcolor{columbiablue}{0.600} & 0.508 & \cellcolor{bisque}{0.565} \\
Toilet &0.180 & 0.180& 0.165 &0.267 &0.292 & \cellcolor{bisque}{0.701} & \cellcolor{columbiablue}0.705 &{0.569} \\
USB & 0.524 & 0.441 & \sbest{0.679} & 0.252 & \cellcolor{bisque}0.654 & 0.561 & 0.575 &\cellcolor{columbiablue}{0.790} \\
WashingMachine &0.006 & 0.241& 0.273 & 0.101& 0.311& \cellcolor{columbiablue}{0.535} & 0.487 &\cellcolor{bisque}{0.526} \\
Window & 0.263 & 0.392 & \cellcolor{columbiablue}{0.833} & 0.668& \cellcolor{bisque}0.766 & {0.754} & 0.728  &0.739 \\
\hline
Avg. (28) & 0.250 &0.313 & 0.457 & 0.259& 0.470& {0.625} & \cellcolor{bisque}0.642 &\cellcolor{columbiablue}{0.665} \\
\hline
Avg. (45) & 0.365 & 0.384& 0.502 & 0.235& 0.451& {0.603} & \cellcolor{bisque}0.615 &\cellcolor{columbiablue}{0.642}\\
\hline
\end{tabular}
\label{tab:partnete_few_sup2}

}
\endgroup

\end{center}

\end{table*}

\begin{figure}[t]
\centering
	\includegraphics[width=\linewidth]{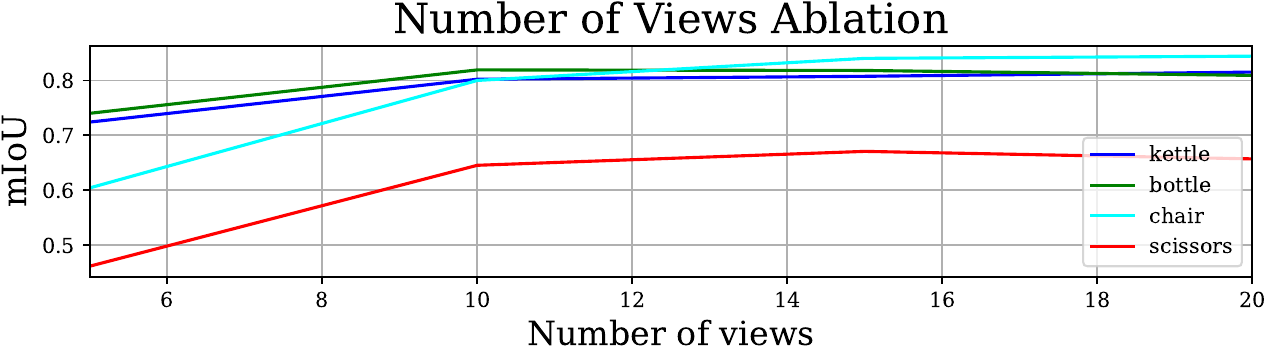}
	\caption{Number of views ablation on PartNetE.}
	\label{fig:view_ablation}
\end{figure}

\subsection{Zero-shot on PartNetE}
\begin{figure*}[t]
\centering
	\includegraphics[width=\linewidth]{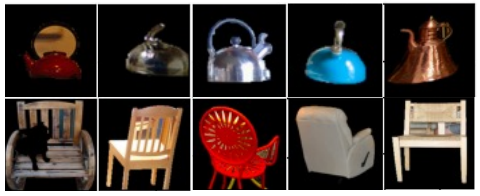}
	\caption{Processed images from PACO~\cite{ramanathan2023paco} that are used as the database for our zero-shot experiment.}
	\label{fig:paco}
\end{figure*}

We present the performance on 18 categories of PartNetE~\cite{liu2023partslip} that overlaps with PACO~\cite{ramanathan2023paco} in Table~\ref{tab:partnete_zero_sup}. With each category, we form a database of at most 150 objects from PACO. We first mask and crop each object in PACO using the ground-truth bounding box and segmentation mask. Objects that are either too small or not visible in the image are subsequently filtered out (see Figure~\ref{fig:paco}). We manually convert labels from PACO to PartNetE for evaluation. For parts that exist in PartNetE but do not exist in PACO, we report the performance as 0 IoU.

We compare 3-By-2 with PartSLIP~\cite{liu2023partslip}, VLPart~\cite{sun2023going}-MC and SAMPro3D~\cite{xu2023sampro3d} + OpenMask3D~\cite{takmaz2023openmask3d}. For PartSLIP, we prompt the pre-trained GLIP model with the language inputs without finetuning, following Liu et al.~\cite{liu2023partslip}. VLPart~\cite{sun2023going} is a SOTA 2D part segmentation method that was trained on a combination of various large-scale 2D part datasets. We replace our 2D part segmentation module with a pre-trained VLPart model, retaining the 3D mask-consistency aggregation module as 3-By-2, and term this baseline VLPart-MC. During inference, to guide VLPart effectively, we prompt the model with language inputs as in PartSLIP. SAMPro3D~\cite{xu2023sampro3d} is a SOTA zero-shot instance segmentation method for 3D scenes using SAM at its core. For semantic segmentation evaluation, we integrate SAMPro3D with OpenMask3D~\cite{takmaz2023openmask3d}, an open-vocabulary 3D scene segmentation method.
\begin{table*}[t!]
\begin{center}

\begingroup
\setlength{\tabcolsep}{2.5pt} 
\renewcommand{\arraystretch}{1.2} 
\caption{Zero-shot performance on the subset of PartNetE~\cite{liu2023partslip} that overlaps with PACO~\cite{ramanathan2023paco}. Our method effectively leverages 2D in-the-wild part segmentation dataset to perform 3D part segmentation.}
\scalebox{0.9}{

\begin{tabular}{|l|c|c|c|c|}
\hline
Categories $\downarrow$  & \begin{tabular}{@{}c@{}}SAMPro~\cite{xu2023sampro3d}+ \vspace{-3pt}\\ OpenMask3D~\cite{takmaz2023openmask3d} \end{tabular} & \begin{tabular}{@{}c@{}}VLPart~\cite{sun2023going}- \vspace{-3pt}\\ MC \end{tabular} & \begin{tabular}{@{}c@{}}PartSLIP \vspace{-3pt}\\ \cite{liu2023partslip} \end{tabular} & 3-By-2\\
\hline
Bottle &0.103 & 0.216 &  \cellcolor{bisque}0.763 & \cellcolor{columbiablue}{0.807}\\
Box & 0.325 & 0.313 & \cellcolor{columbiablue}{0.575} & \cellcolor{bisque}0.369 \\
Bucket & 0.035 & \cellcolor{columbiablue}0.253 & 0.020 & \cellcolor{bisque}0.124 \\
Chair & 0.024 & 0.066 & \cellcolor{columbiablue}{0.607} & \cellcolor{bisque}0.468 \\
Clock & 0.007 & 0.205 & \cellcolor{columbiablue}{0.267} & \cellcolor{bisque}0.253 \\
FoldingChair & 0.437 & \cellcolor{bisque}0.813 & \cellcolor{columbiablue}{0.917} & 0.712\\
Kettle & 0.026 & \cellcolor{bisque}0.211 & 0.208 & \cellcolor{columbiablue}{0.765}\\
Knife & 0.267 & \cellcolor{bisque}0.427 & \cellcolor{columbiablue}{0.468} & 0.413 \\
Lamp & 0.074 & 0.166 & \cellcolor{bisque}0.371 & \cellcolor{columbiablue}{0.500} \\
Laptop & 0.017 & 0.060 & \cellcolor{bisque}0.270 & \cellcolor{columbiablue}{0.394} \\
Microwave & 0.000 &\cellcolor{bisque}0.192 & 0.166 & \cellcolor{columbiablue}{0.348}\\
Mouse & 0.019 & 0.000 & \cellcolor{bisque}0.270 & \cellcolor{columbiablue}{0.307} \\
Pen & 0.088 & \cellcolor{bisque}0.016 & \cellcolor{columbiablue}{0.146} & 0.010 \\
Pliers & \cellcolor{columbiablue}0.996 & 0.494 & 0.054 & \cellcolor{bisque}0.944 \\
Remote & 0.084 & \cellcolor{bisque}0.132 & 0.115 & \cellcolor{columbiablue}{0.239} \\
Scissors & 0.118 & 0.193 & \cellcolor{bisque}0.218 & \cellcolor{columbiablue}{0.594} \\
Table & 0.004 & 0.114 & \cellcolor{columbiablue}{0.398} & \cellcolor{bisque}0.251 \\
TrashCan & 0.000 & 0.119 & \cellcolor{columbiablue}{0.301} & \cellcolor{bisque}0.239 \\
\hline
Avg. (18) & 0.146 & 0.222 & \cellcolor{bisque}0.341 & \cellcolor{columbiablue}{0.430} \\
\hline
\end{tabular}
\label{tab:partnete_zero_sup}

}
\endgroup

\end{center}

\end{table*}

\begin{figure*}[t]
\centering
	\includegraphics[width=\linewidth]{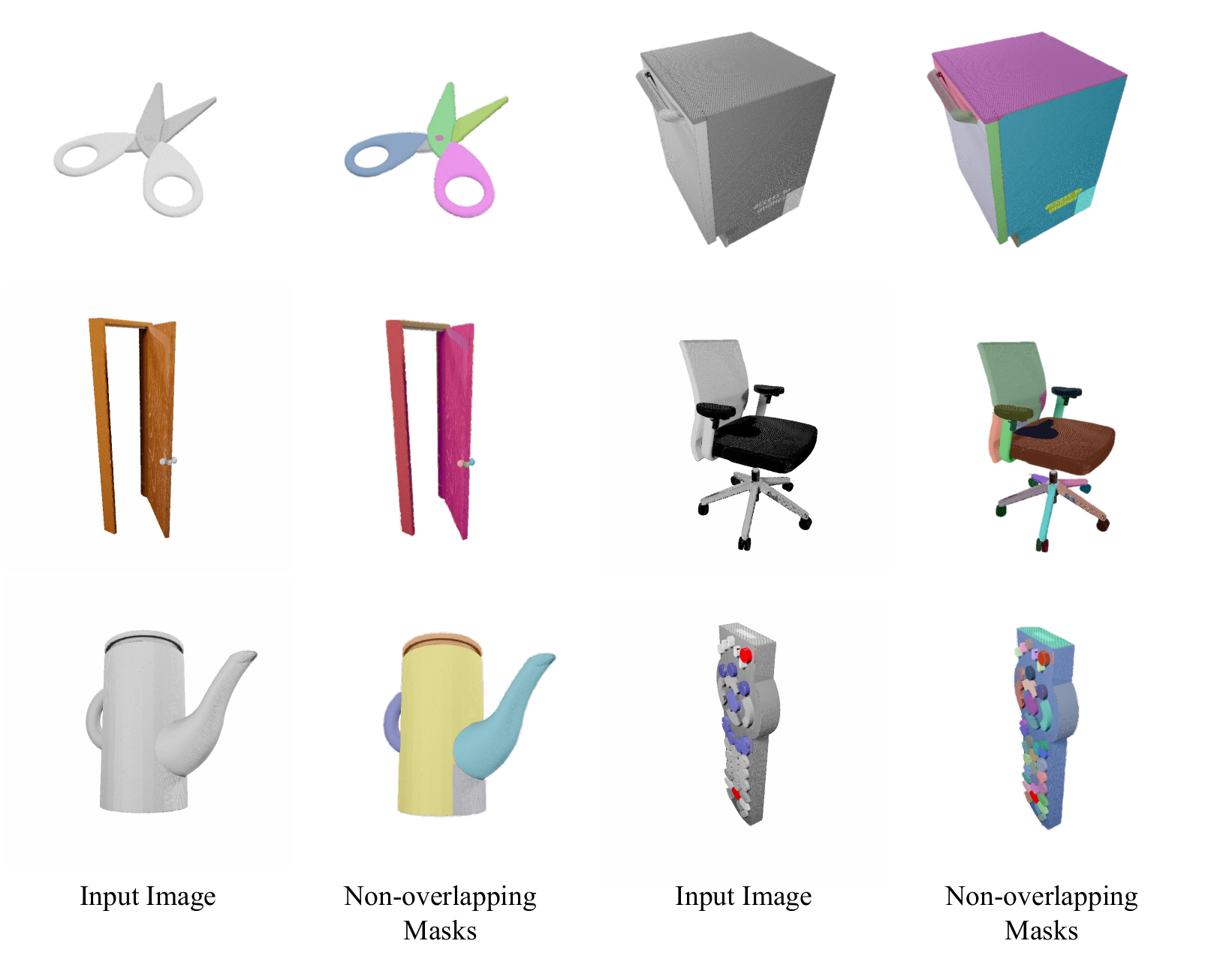}
	\caption{Non-overlapping masks generated by our method.}
	\label{fig:non_overlapping_masks_sup}
\end{figure*}

\section{More Ablation Results}\label{sec_sup:ablation}
\subsection{Non-overlapping Mask Generation} In Figure~\ref{fig:non_overlapping_masks_sup}, we demonstrate the visualization of non-overlapping mask outputs from our non-overlapping mask generation module.

\begin{figure*}[t]
\centering
	\includegraphics[width=\linewidth]{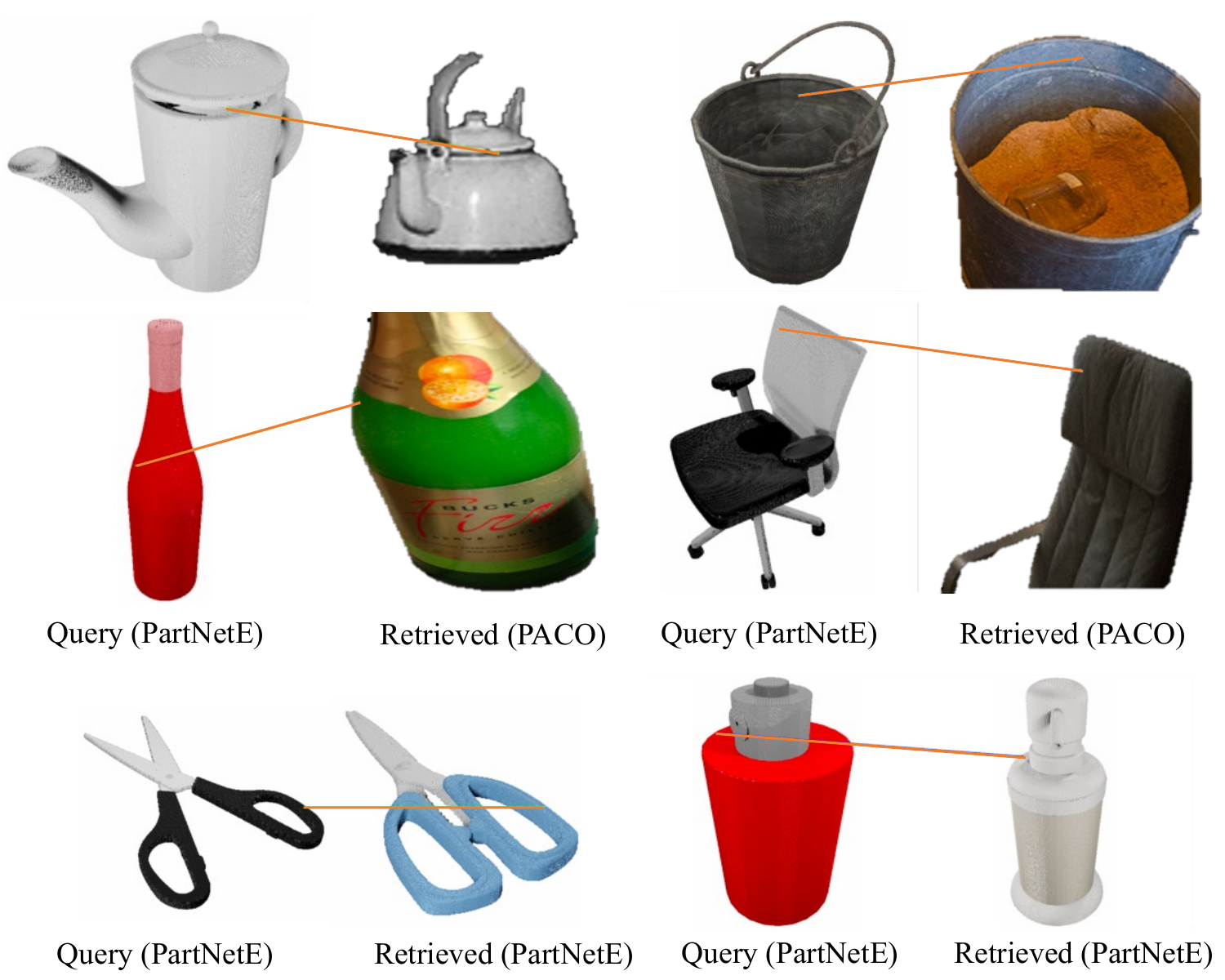}
	\caption{(Top) Correspondence retrieval between PartNetE test objects and PACO. (Bottom) Correspondence retrieval between PartNetE few-shot and test sets. Our method using Diffusion Features~\cite{tang2023emergent} perform well in both cases.}
	\label{fig:corr_zero}
\end{figure*}

\subsection{Results on Multiple Runs} Table~\ref{tab:multi_sup} presents the few-shot setting results across multiple runs. For each category, we conduct label transferring independently 3 times and report the average mIoU along with the standard deviation across these runs. Note that the standard deviation demonstrates a relatively small variation between the runs, indicating that sparsely sampling points inside the mask for label transferring does not have a significant impact on the final performance of our model.

\begin{table}[t!]
\begin{center}

\begingroup
\setlength{\tabcolsep}{3pt} 
\renewcommand{\arraystretch}{1.2} 
\caption{Average mIoU and standard deviation for 3 independent runs. Randomly sampling pixels for label transfer has minimal effect on the final performance of the method.}
\vspace{-5pt}
\scalebox{0.9}{

\begin{tabular}{|cc|cc|cc|cc|}
\hline
\multicolumn{2}{|c|}{Scissors} & \multicolumn{2}{c|}{Kettle} & \multicolumn{2}{c|}{Suitcase} & \multicolumn{2}{c|}{Bottle}\\
\hline
Avg. & Std. & Avg. & Std. & Avg. & Std. & Avg. & Std.\\
\hline
0.675 & 0.008 & 0.663 & 0.002 & 0.813 & 0.003 & 0.810 & 0.007\\
\hline
\end{tabular}

\label{tab:multi_sup}
}
\endgroup

\end{center}

\end{table}

\begin{table}[t!]
\begin{center}

\begingroup
\setlength{\tabcolsep}{3pt} 
\renewcommand{\arraystretch}{1.2} 
\caption{Ablation for different number of instances in the few-shot database on Kettle. We demonstrate strong performance even with only 1 labeled instance.}
\vspace{-5pt}
\scalebox{1}{

\begin{tabular}{|l|ccc|c|c|}
\hline

Database & Lid & Handle & Spout & Avg. & PartSLIP Avg.\\
\hline
1-instance & 0.762 & 0.901 & 0.765 & 0.809 & 0.200\\
4-instance & 0.784 & 0.893 & 0.774 & 0.817 & 0.690\\
8-instance & 0.759 & 0.904 & 0.783 & 0.815 & 0.770\\
\hline
\end{tabular}
\label{tab:instances}
}
\endgroup

\end{center}

\end{table}

\subsection{Number of Views Ablation}
In Fig.~\ref{fig:view_ablation}, we ablate the number of views. Using more than 10 views has minimal impact on performance, aligning with the observations in PartSLIP~\cite{liu2023partslip}. In all experiments, we follow the optimal number of views for all baselnes.

\subsection{More Analyses on Few-shot Database}
Table~\ref{tab:instances} presents the results for Kettle using various numbers of object instances in the database for few-shot label transfer. The instances are randomly selected from the few-shot set of 8 objects. Note that even with a single instance in the database, 3-By-2 achieves remarkable performance, significantly outperforming PartSLIP~\cite{liu2023partslip}. The performance gap between the 1-instance and 8-instance database settings for our method is negligible, demonstrating that 3-By-2 can effectively operate with a minimal number of labeled objects.

\section{Implementation Details}\label{sec_sup:implementation}
We use the image diffusion feature extractor~\cite{tang2023emergent} as our backbone. We extract features at the first upsampling block of the U-Net at time $t=261$. We use an empty string as prompt input to the image diffusion model. For SAM~\cite{kirillov2023segment}, we use point sampling as the prompt with a grid size of $64\times 64$. For all majority voting operations, we employ a cut-off threshold of $0.6$. That is, if $x_i \geq 0.6~\Sigma_j x_j$ and $x_i=\max_j x_j$ then $x_i$ is the dominant element of $\{x_j\}_{j=1,\dots,N}$.

\section{Semantic Correspondence Matching Visualization}\label{sec_sup:viz}
We show the visualization of the semantic correspondence matching in Figure~\ref{fig:corr_zero}. For each pixel in the query object, we find the closest match from the database using cosine distances of the features extracted from DIFT~\cite{tang2023emergent}. The semantic matches are high quality for both PACO (zero-shot) and PartNetE (few-shot).

\section{Limitations \& Future Work}\label{sec_sup:discussion}
While our approach offers flexibility across multiple levels of granularity and diverse object categories, it is important to acknowledge certain limitations. The performance of our method is directly tied to the performance of SAM~\cite{kirillov2023segment} and DIFT~\cite{tang2023emergent}. 
Additionally, for scenarios like animal or human part segmentation, where part boundaries are usually not determined by edges, applying SAM in this case may prove suboptimal for 2D mask proposal. However, we note that one can effortlessly switch out SAM and DIFT for other segmentation and correspondence matching methods. The choice for leveraging these models stems from their status as current SOTA approaches. Future improvements of either one of these tasks can further improve the performance of our approach. In addition, noise in labels in the database can lead to incorrect label assignments, even when the correspondences are accurately identified by the DIFT features.

Due to the strong taxonomy preservation capability of our method (see Sec.~\ref{sec:ablation}), future work can explore applying similar approach to other 3D tasks such as object classification or scene semantic segmentation.

\section{Additional Qualitative Results}\label{sec_sup:qualitative} 
\begin{figure*}[t]
\centering
	\includegraphics[width=\linewidth]{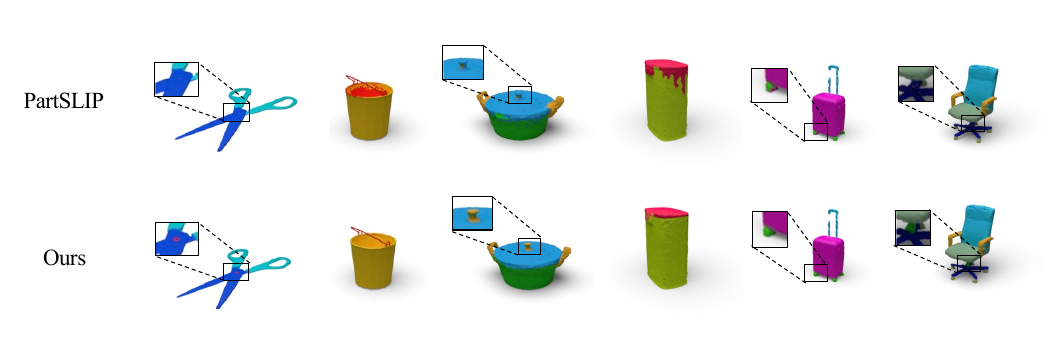}
	\caption{Few-shot qualitative comparison between PartSLIP~\cite{liu2023partslip} and 3-By-2. Our approach is more successful at capturing small parts and details.}
	\label{fig:sup_compare}
\end{figure*}

\begin{figure*}[h]
 \vspace{-10pt}
\centering
	\includegraphics[width=\linewidth]{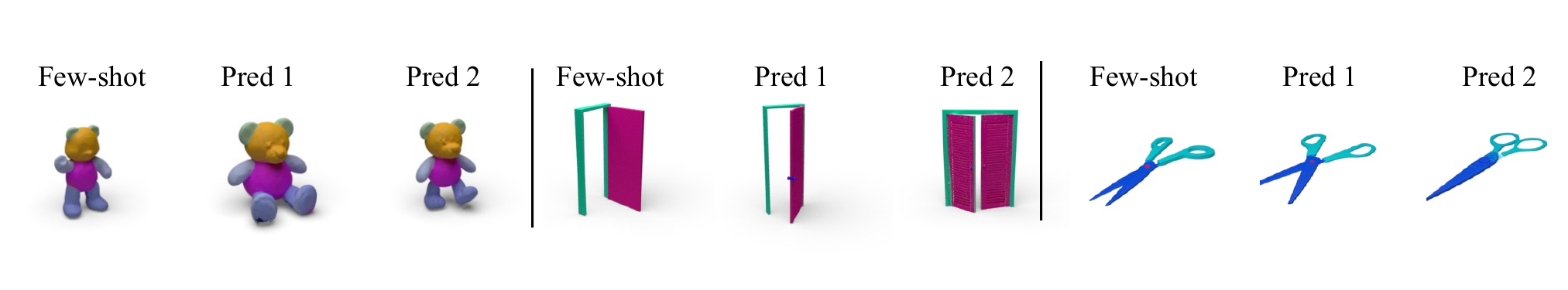}
  \vspace{-20pt}
	\caption{Few-shot is the ground-truth segmentation template, Pred1 and Pred2 are the segmentation predictions on 2 different samples given the few-shot template masks.}
 \vspace{-10pt}
	\label{fig:rebuttal_teddy}
\end{figure*}

In Fig.\ref{fig:sup_compare}, we present a qualitative comparison between PartSLIP\cite{liu2023partslip} and our approach. Our method, 3-By-2, demonstrates a superior ability to capture small parts with higher precision. We further demonstrate the qualitative performance of our method on non-rigid objects, including deformable (teddy bear) and articulated (door and scissors) objects (Fig.~\ref{fig:rebuttal_teddy}).

\bibliographystyle{splncs04}
\bibliography{egbib}
\end{document}